\pdfoutput=1

\documentclass[11pt]{article}
\usepackage{setspace}
\usepackage[dvipsnames]{xcolor}
\usepackage{graphicx}
\usepackage{tabularx}
\usepackage{afterpage}
\usepackage{booktabs}
\usepackage{amssymb,amsmath}
\usepackage{subcaption}
\usepackage{xspace}
\usepackage{etoolbox}
\usepackage{enumitem} %
\usepackage{cellspace} %
\usepackage{listings}
\usepackage{enumitem} %
\usepackage{multirow} %
\usepackage[]{acl} 
\providetoggle{showcomments}
\toggletrue{showcomments}

\usepackage{times}
\usepackage{latexsym}
\usepackage{xspace}
\usepackage[T1]{fontenc}

\usepackage[utf8]{inputenc}

\usepackage{microtype}
\usepackage{inconsolata}
\usepackage{float}

\usepackage[utf8]{inputenc}
\usepackage{textgreek} %
\title{
A Similarity Measure for Comparing Conversational Dynamics
}

\author{
\hspace{-1.1cm}
  Sang Min Jung\thanks{\ Equal contribution.} \\
\hspace{-1.1cm}
  Cornell University \\
\hspace{-1.1cm}
  \texttt{sj597@cornell.edu}
  \And
\hspace{-2cm}
  Kaixiang Zhang\footnotemark[1] \\
\hspace{-2cm}
  Cornell University \\
\hspace{-2cm}
  \texttt{kz88@cornell.edu}
  \And
\hspace{-1.2cm}
  Cristian Danescu-Niculescu-Mizil \\
\hspace{-1.2cm}
  Cornell University \\
\hspace{-1.2cm}
  \texttt{cristian@cs.cornell.edu}
}

\newcommand{\bluetext}[1]{\textcolor{blue}{#1}}
\newcommand{\redtext}[1]{\textcolor{red}{#1}}
\newcommand{\revision}[1]{#1}

\newcommand{\cut}[1]{}
\newcommand{\xhdr}[1]{{\noindent\bfseries #1.}}

\newcommand{\definedas}{\overset{\triangle}{=}}
\newcommand{\measurename}{ConDynS\xspace}
\newcommand{\dynamics}{dynamics\xspace}
\newcommand{\Dynamics}{Dynamics\xspace}

\newcommand{\ppatterns}{patterns\xspace}

\newcommand{\anchor}{anchor\xspace}

\newcommand{\groupone}{Cluster~1\xspace}
\newcommand{\grouptwo}{Cluster~2\xspace}
\newcommand{\groupdelta}{set $\Delta$\xspace}
\newcommand{\groupnodelta}{set $\neg \Delta$\xspace}

\newcommand{\sop}{SoP\xspace}

\begin{document}

\maketitle

\begin{abstract}

The quality of a conversation goes beyond the individual quality of each reply, and instead emerges from how these combine into interactional 
dynamics
that give the conversation its distinctive overall ``shape''.
However, there is no robust automated method for comparing conversations in terms of their overall 
dynamics.
Such methods could enhance the analysis of conversational data and help evaluate conversational agents more holistically.

In this work, we introduce a similarity measure for comparing conversations with respect to their dynamics.
We design a validation 
procedure
for testing the robustness of the metric in capturing differences in conversation dynamics and for assessing its sensitivity to the topic of the conversations.
To illustrate the measure's utility, we use it to analyze conversational dynamics in a large online community, bringing new insights into the role of situational power in conversations.

\end{abstract}

\section{Introduction}
\label{sec:intro}

In a conversation, individual utterances combine to form interactional patterns, such as  
changes in tone (e.g., from passive-aggressive to defusing), 
conversational strategies (e.g., analogies, concessions, or challenges), and
interaction sequences (e.g.,  extended back-and-forth vs. one-sided rants).
Each of these patterns contributes to shaping the conversation's overall \textit{dynamics}, but none of them alone is sufficient to characterize it \cite{tannen_conversational_2005,hua_how_2024}.

These emerging conversational dynamics are closely tied to the perceived quality of the conversation and its outcome \cite[inter alia]{stasi_zooming_2023, dcosta_what_2024, liao_conversational_2023}.
As such, a measure comparing conversations with respect to their  overall 
dynamics can enhance our ability to analyze human-human and human-AI conversational data.  
For example, it can be used to group conversations according to their dynamics and distinguish those that are likely to lead to positive outcomes.
This type of analysis could enable a more holistic evaluation of conversational agents, one that goes beyond optimizing for the quality of each response to encourage overall dynamics that are desirable.

However, developing a method for comparing conversations with respect to their overall dynamics presents several challenges.
The first 
challenge 
is finding an appropriate way of representing the dynamics of a conversation: it is not sufficient to detect individual patterns separately (e.g., speech acts, empathy, politeness, sarcasm), as done by prior work  \cite{ghosh_role_2017, oraby_are_2017, chhaya_frustrated_2018,danescu-niculescu-mizil_computational_2013}.
Instead, a representation of the overall dynamics must capture 
how
relevant patterns 
of different types connect to each other.
For example, a passive-aggressive tone changing into a defusing tone leads to a very different dynamic than when a defusing tone is followed by a passive-aggressive tone.

The second challenge arises when comparing dynamics.
Dynamics take place at multiple scales, with some patterns spanning single exchanges (e.g., a sarcastic response) and others spanning the entire conversation (an increasingly escalating tone).
Furthermore, 
a single utterance
can contribute to multiple patterns (e.g., an 
utterance
can be a sarcastic response and simultaneously be part of an increasingly escalating tone).
This inherent overlap makes it hard to align the dynamics of two conversations in order to quantify how similar they are.

In this work, we address these challenges to introduce 
a
similarity measure for conversational dynamics: \measurename (read as ``condense'').
We address the first challenge by representing dynamics as a sequence of relevant interactional patterns in a conversation (a sequence of patterns, henceforth the \textit{SoP}), extracted from 
a summary
of conversational dynamics \cite{hua_how_2024}.
This representation captures not only which interaction patterns are present in a conversation, but also the order in which they follow each other.

\begin{figure*}[ht] 
    \centering
    \includegraphics[width=1\textwidth]{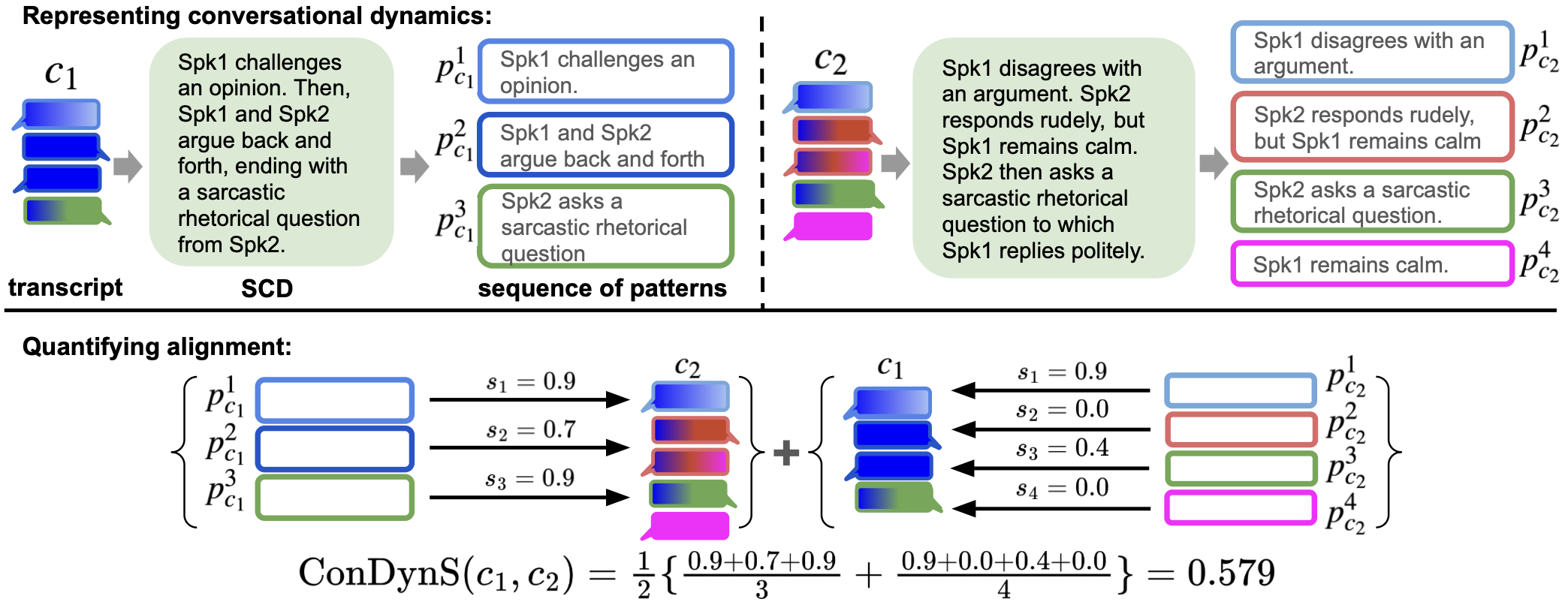}
    \caption{Representing dynamics and quantifying their alignment to calculate \measurename. Colors represent interactional patterns, sometimes spanning multiple utterances; also, an utterance can contribute to multiple patterns.
    }
    \label{fig:our_measure}
\end{figure*}

We address the second challenge by designing an asymmetric procedure for aligning conversational dynamics (\autoref{fig:our_measure}).
The main intuition behind this procedure is to combine the advantage of the SoP representation---which allows checking the order in which interaction patterns appear---with the advantage of a simple transcript representation---in which we can find patterns with high-recall,
even when they are overlapping.

To validate the effectiveness of \measurename and compare it with baseline measures using other representations or alignment methods, we introduce a human-in-the-loop procedure for generating labeled data. 
\measurename recovers these labels with over 90\% accuracy, substantially outperforming the baselines, while being robust against 
topical confounds.

We further demonstrate how a similarity measure for conversational dynamics can enable 
new types
of analysis
\revision{by applying \measurename to conversations from 
a large online debate community.}
First, it allows us to adapt standard similarity-based 
techniques---clustering, inter-group similarity, and intra-group diversity---to study conversational dynamics.
Second, we use our measure to investigate which participants are more likely to influence the dynamics of a conversation, 
providing new insights into the role of situational power in conversations.

In summary, 
in this work we: 
\begin{itemize}
    \item introduce a similarity measure for comparing conversational dynamics;
    \item propose a validation procedure that enables comparison against baseline measures;
    \item use our measure to provide new insights into the role of situational power in conversations. 
\end{itemize}

\revision{
We additionally explore the versatility of our measure by applying it to two other conversational domains, including scripted casual conversations between friends and non-English discussions held in a collaborative setting.
To encourage further use and development, we release the code for \measurename publicly as part of ConvoKit, including demos on multiple datasets.\footnote{\url{https://convokit.cornell.edu}}}

\section{Background and Related Work}
\label{sec:related_work}

\xhdr{Conversational dynamics}
\revision{
We use the term ``conversational dynamics'' to denote how different interactional patterns combine and unfold over the course of a conversation to shape its overall temporal trajectory.
This is arguably a broad and under-defined concept, as it depends on what type of interactional patterns are considered most salient in a given context, and how these patterns are identified and tracked over the course of the interaction.
Prior literature has studied it by focusing on the temporal sequence of specific utterance-level patterns in isolation,
such as argumentation strategies \cite{morio_revealing_2019, mirzakhmedova_unveiling_2023}, sentiment \cite{wang_piece_2014}, communicative acts \cite{liao_conversational_2023},
or dialog acts \cite{zhao_floweval_2022}.
Other work has focused on tracking patterns that only emerge at a higher multi-turn or conversational-level, such as turn-taking \cite{sacks_simplest_1974}, 
time-sharing \cite{zhang_time_2025}, 
coordination \cite{fusaroli_investigating_2016}, 
or changes in topical entropy \cite{fischer_personality_2024}.}

\revision{
In this work, we follow \cite{hua_how_2024} to take a holistic approach that does not limit the types of patterns that can contribute to the overall dynamics.
By adopting this flexible framework, our measure can compare dynamics involving patterns that are potentially novel or domain-specific, rather than imposing strict theoretical constraints in advance. 
}

\xhdr{Measuring conversation-level similarity}
Prior work on measuring conversation-level, rather than utterance-level, similarity is limited.
\citet{lavi_weve_2021} adapts edit distance to measure similarity of ``dialogue flow'', by defining substitution cost based on the semantic similarity of utterances. 
Other methods \cite{bhaumik_taskdiff_2023} additionally consider semantic features specific to task-oriented interactions, such as agent intent.
In contrast, \measurename is not concerned with the topic or semantics of what is discussed, focusing solely on interactional patterns and the emerging dynamics.

Other measures 
focus 
on a few predefined features, such as dialog acts \cite{enayet_analysis_2022, zhao_floweval_2022}, sentiment \cite{xu_clustering-based_2019}, and number of words per turn \cite{appel_combining_2018}.
Our measure instead compares dynamics that emerge from how \textit{multiple} types of interactional patterns combine and unfold.

\xhdr{Synthetic conversations}
LLMs have been used to generate and annotate datasets across NLP tasks (see \citet{tan_large_2024} for a survey) including in the conversational domain \cite{wang_patient-_2024, louie_roleplay-doh_2024,liu_personality-aware_2024},
sometimes with expert human input \cite{louie_roleplay-doh_2024}.
We build on this work to design our validation procedure, which uses human-written summaries to generate conversations with labels for relative similarity.
\revision{We use real conversations for the rest of our analysis.}

\section{Measure}
\label{sec:our_measure}
Measuring the similarity between the dynamics of two conversations involves (1) representing these dynamics and (2) comparing them.
Below, we discuss several options 
for these steps,
which combine to form \measurename and several baseline measures.
Here we describe the general approach, and defer to Section \ref{sec:datatset} for details about the operationalization in the specific domains we analyze in this work.

\subsection{Representing conversational dynamics}
Conversational dynamics are complex, emerging from the progression and juxtapositions of multiple interaction \ppatterns.
Therefore, their representation 
must go beyond 
describing
individual patterns separately (e.g., how polite each reply is, whether it is sarcastic or not, etc.), and instead capture how relevant patterns combine to form the conversation's dynamics.
Given that patterns are often overlapping and can emerge at multiple scales---some being confined to a single utterance while others are spanning multiple utterances---there is an inherent tradeoff between precisely representing a coherent progression and capturing all patterns present in a conversation.

At one extreme, the \textbf{raw transcript} offers the most comprehensive representation of a text-based conversation.
By preserving all information, it implicitly includes all the patterns that combine to form its conversational dynamics.
However, this is a noisy representation as the patterns are not explicitly identified, nor are they separated from the topical context in which they appear. 
This noise is problematic for our purposes as it might interfere with comparisons focused solely on conversational dynamics.
Furthermore, it lacks an explicit ordering of the patterns, making it hard to compare the progression of the interaction.

The \textbf{summary of a conversation's dynamics} (or \textit{SCD}) offers an alternative representation that abstracts away the topical content and explicitly identifies interactional patterns \cite{hua_how_2024}.
Through their abstraction, SCDs 
select a subset of the interactional patterns that are deemed most relevant to the overall conversation's trajectory.
SCDs thus offer a more condensed and precise representation than the transcripts.
This, however, necessarily comes at the expense of recall.

To explicitly capture the order in which individual interactional patterns occur, 
an SCD
can be structured into a \textbf{sequence of patterns} (\sop).  
These are 
ordered lists
of natural language strings extracted from SCDs, 
each representing one pattern.
\autoref{fig:our_measure} (top) illustrates the steps of obtaining a SoP from the raw transcript of a conversation, and full examples from our dataset are included in \autoref{appendix:examples_scd_sop}.
The exact operationalization of each step is dependent on the application domain, and is detailed in Section \ref{sec:datatset}.

\subsection{Comparing dynamics}
A straightforward approach to 
compare dynamics
would be
 to measure how well the interaction patterns in one conversation \textit{match}
with the ones in the other conversation.
Our approach 
additionally
recognizes the role of the order in which patterns appear and quantifies how well the patterns in the two conversations are \textit{aligned}.

\xhdr{Matching: baselines} 
To form our baselines, we 
apply existing text-similarity metrics to quantify how well dynamics match across two conversations:
\begin{itemize}[itemsep=0pt, leftmargin=*]
    \item Cosine similarity of SBERT embeddings: Using a SBERT sentence transformer \cite{reimers_sentence-bert_2019}, we calculate the cosine similarity of the two conversations.
    \item BERTScore: We use BERTScore \cite{zhang_bertscore_2020} to compare the similarity of the two conversations.
    \item Naive prompting: We prompt a large language model to compare two conversations in terms of their \dynamics and give a similarity score between $1$ and $100$.
    The prompt is in the Appendix in Figures \ref{fig:naive_prompt} and \ref{fig:naive_prompt_scd}.
\end{itemize}
All these metrics can be applied to either the transcript representation or the SCD representation, resulting in six baseline measures.  

\xhdr{Alignment: \measurename} While straightforward, these matching metrics ignore the order in which the interaction patterns follow each other to give rise to the overall dynamics. 
We address that by designing a new metric that quantifies how well the \textit{sequence} of patterns in one conversation aligns with the dynamics of another conversation.

\newcommand{\cone}{c_1}
\newcommand{\ctwo}{c_2}

Formally, let $P_{\cone} = [p^1_{\cone}, p^2_{\cone}, \ldots, p^n_{\cone}]$ denote the SoP of conversation $\cone$.  
Let $\ctwo$ be another conversation with whose dynamics we want to compare; we purposefully defer the discussion of the representation of $c_2$.
We define an alignment vector
\begin{equation}
   s(P_{\cone}, \ctwo) = [s_1, s_2, \ldots, s_n] \in [0,1]^n,
\end{equation}

\noindent where $s_i \in [0,1]$ indicates how much $p^i_{\cone} \in P_{\cone}$ contributes to the alignment with the dynamics of $\ctwo$.
In addition to rewarding patterns that also appear in $\ctwo$, the score is designed to penalize %
patterns that: (1) appear out of order in $\ctwo$, and (2) are separated in $\ctwo$ from the previous pattern in the $\cone$ sequence (e.g., by other patterns that only appear in $\ctwo$).
At the extremes, a pattern $p^i_{\cone}$ that does not appear in $\ctwo$ will receive a score $s_i = 0$ and a pattern $p^i_{\cone}$ that also appears in $\ctwo$ immediately after a pattern matching $p^{i-1}_{\cone}$ will have a score $s_i = 1$.  

We average these scores to quantify how well $\cone$'s sequence of patterns aligns with those in $\ctwo$:
\begin{equation}\label{eq:asym}
(\cone \rightarrow \ctwo) \definedas \frac{1}{|P_{\cone}|} \sum_{s_i \in s(P_{\cone}, \ctwo)} s_i.   
\end{equation}
\noindent We note that this is an asymmetric measure, and that we can analogously compute $(\ctwo \rightarrow \cone)$, i.e., how well $\ctwo$'s sequence of patterns aligns with those in $\cone$.\footnote{As an extreme example that renders this asymmetry  evident, consider a hypothetical case in which $\cone$ is a conversation starting with all the utterances of $\ctwo$ and continuing with more replies. While $\ctwo$'s sequence of patterns will align perfectly with $\cone$, the sequence in $\cone$ will not align perfectly due to patterns appearing only in the continuation.}   
We average these two asymmetric scores to obtain our similarity measure:
\begin{equation}
\text{\measurename}(\cone, \ctwo) \definedas \frac{1}{2}\{(\cone \rightarrow \ctwo) + (\ctwo \rightarrow \cone)\}.
\end{equation}

In terms of representation, in Eq. (\ref{eq:asym}) $\cone$ is represented as a SoP to account for the order in which the patterns appear.  
However, given its asymmetry, we have a choice of how to represent $\ctwo$ when calculating the alignment vector $s(P_{\cone}, \ctwo)$.
One option is to also use the SoP representation to focus on the most relevant patterns and exploit their explicit ordering.
However, since our goal 
at this step
is to check for the presence of a specific pattern in $\ctwo$, 
recall is especially important.
As such, we propose using the most comprehensive representation of $\ctwo$: its raw transcript. 
This way, the asymmetric nature of the alignment procedure allows us to combine the precision and ordering of the SoP representation with the recall of the transcript representation.

\section{Data and Operationalization}
\label{sec:datatset}
\xhdr{Online debate discussions}
To validate and demonstrate applications of \measurename, we use a dataset of conversations from the ChangeMyView subreddit (CMV), retrieved from ConvoKit \cite{chang_convokit_2020}.
The objective of this platform is for participants 
(Challengers)
to persuade the original poster (OP) to change their viewpoint on 
an opinion they hold.\footnote{We follow prior work and consider a conversation to be one linear reply-chain starting with the first comment to the original post introducing the to-be-changed opinion \cite{chang_trouble_2019, hua_how_2024}.}
The dataset includes conversations from the subreddit's inception 
in 2015, up to 2018, and is thus not polluted by content generated by large language models.
In this paper we use a total of 9,138 CMV conversations, selected as described in Sections \ref{sec:validation} and \ref{sec:application}.

This setting has several properties that make it particularly suitable for developing a similarity metric for conversational dynamics.
First, it has been a resource for many studies analyzing how conversational features connect to different outcomes---such as successful persuasion \cite{tan_winning_2016, priniski_attitude_2018, monti_language_2022, wei_is_2016} or conversation derailment \cite{altarawneh_conversation_2023, kementchedjhieva_dynamic_2021, chang_trouble_2019}---documenting its richness in conversational dynamics.
Second, a key feature of the dataset is the “delta” ($\Delta$) mechanism through which the OP can award a $\Delta$ to 
a Challenger
that successfully changed their view.
This mechanism provides explicit persuasion labels for each conversation, which we will use to interpret our results.
Finally, \citet{hua_how_2024} developed the SCDs procedure on this dataset. 
As such, they distribute human-written SCDs and provide a validated procedural prompt for automatically generating SCDs,
which grounds our
method and validation procedure in an established framework.

\xhdr{Other conversation settings}
\revision{
As discussed in Section \ref{sec:related_work}, we adopt a flexible framework for conversational dynamics in order to allow adapting the measure to other domains where different patterns might be at play.
To explore the versatility of \measurename, we apply it to two additional settings (Section \ref{sec:application2}). 
The first setting involves 50 fictional conversations from the \textit{Friends} TV show \cite{chen_character_2016} which are scripted to resemble everyday face-to-face interactions and reflect entertainment-driven dialogue.
The second is a collaborative setting in a non-English language, specifically 100 conversations from the German Wikipedia talk-pages \cite{hua_wikiconv_2018}.
}

\xhdr{Operationalization} 
\revision{
We release a modular implementation of \measurename in ConvoKit, making it easy to swap specific components to facilitate adaptation to different settings.
We use Google's Gemini 2.0 Flash model's API \cite{anil_gemini_2024} for generating SCDs, extracting \sop, and quantifying the alignment of dynamics.\footnote{We also experimented with OpenAI's chatgpt-4o model \cite{achiam_gpt-4_2024} on the validation set, without noting substantial changes in performance (Appendix \ref{appendix:additional_val}).}
To generate SCDs, we use the procedural prompt validated by \citet{hua_how_2024} for our main CMV setting, and modified versions that include domain-specific examples for the additional settings. 
All generated SCDs are distributed together with the respective datasets in ConvoKit. 
To measure the alignment scores $s_i$, we use a few-shot in-context learning prompt with human-constructed examples (showing scoring and reasoning) to quantify alignment.
All prompts are included in \autoref{appendix:ourmeasure_prompts}.
}

\section{Validation}
\label{sec:validation}

\begin{figure}[t!] 
    \centering
    \includegraphics[width=0.45\textwidth]{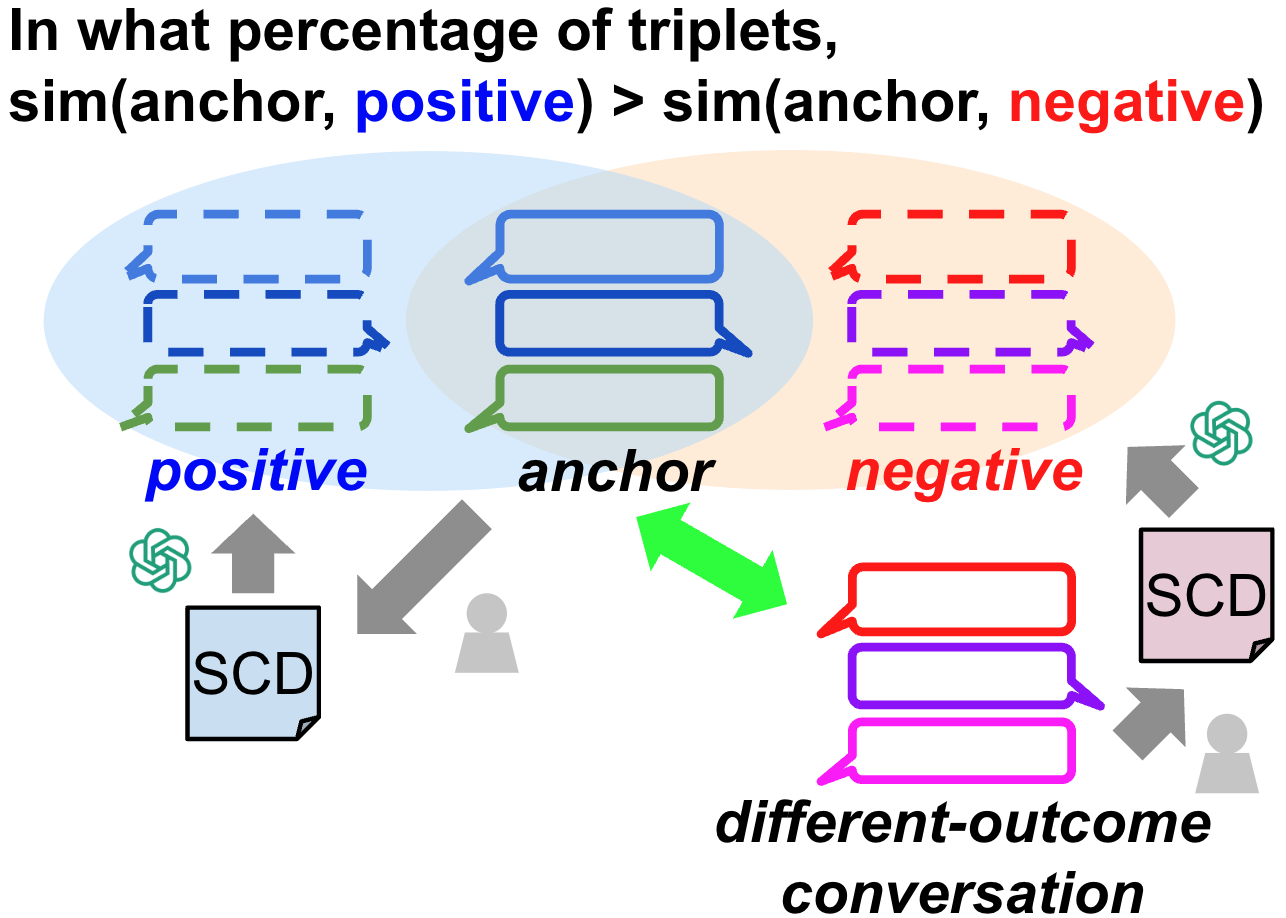}
    \caption{Overview of the validation procedure. Simulated conversations are shown with dashed lines.}
    \label{fig:validation_setup}
\end{figure}

\begin{table*}[t!]
    \centering
    \begin{tabular}{p{2.2cm}p{1.5cm}p{1cm}p{1cm}p{1cm}p{1cm}p{1cm}p{1cm}p{1cm}}
    \toprule
    {Measure} & \multicolumn{2}{c}{\measurename} & 
    \multicolumn{2}{c}{cosine sim.} & \multicolumn{2}{c}{BERTScore} & \multicolumn{2}{c}{Naive prompting}\\
    {Representation} & {SoP+Trx} & {SoP} & { Trx} & { SCD} & { Trx} & { SCD} & { Trx} & { SCD}\\
    \midrule
    { same topic} & { \textbf{92}\%} & { 86\%} & { 52\%} & { 66\%} & { 62\%} & { 72\%} & { 58\%} & { 80\%}\\
    { different topic} & { \textbf{94}\%} & { 80\%} & { 50\%} & { 74\%} & { 56\%} & { 72\%} & { 68\%} & { 72\%}\\
    { adversarial} & { \textbf{86}\%} & { 84\%} & { 2\%} & { 66\%} & { 10\%} & { 70\%} & { 44\%} & { 56\%}\\
    \bottomrule
    \end{tabular}
    \caption{\label{citation-guide} Accuracy of each similarity metric in our validation 
    experiment,
    for different topic conditions.
    Each baseline is either given the raw transcript (abbreviated as \textit{Trx} above) as the input or the raw machine-generated \textit{SCD}. The highest score for each topic condition is bolded.}
    \label{tab:result_table}
\end{table*}

No data with labels for similarity of conversational dynamics is available, and the vast space of possible dynamics and their complexity makes human-annotation highly subjective and prohibitively time-consuming \cite{xu_clustering-based_2019,lavi_weve_2021}.
Therefore, to validate our measure and compare it with baseline measures, we design a human-in-the-loop procedure for obtaining 
synthetic
data in which the relative similarity of conversational dynamics is known (Figure~\ref{fig:validation_setup}).

Specifically, we construct triplets of conversations with 
(1) an \textit{\anchor} conversation that serves as the reference for comparison, 
(2) a \textit{positive} conversation with a dynamic that is known to be similar to that of the \anchor,
and (3) a \textit{negative} conversation with a dynamic that is known to be different from that of the \anchor.\footnote{The anchor/positive/negative terminology is not related to sentiment, and is borrowed from \citep{schroff_facenet_2015}.}
Given a collection of such triplets, we calculate the accuracy of a similarity measure 
as
the proportion of triplets where the anchor-positive pair receives a higher similarity score than the anchor-negative pair.

\xhdr{Anchor-positive pairs}
 Recent work has demonstrated that LLMs can be used to reliably simulate conversations with specific properties \cite{wang_patient-_2024, liu_personality-aware_2024}.
We use a similar idea and prompt an LLM to simulate a conversation that closely follows the \dynamics of a given anchor conversation.
A manual check of the resulting pairs, however, reveals that 
directly providing the anchor's transcript in the prompt
often leads the model to directly replicate its surface-level features, 
such as topic, word choice, or speaker turn order, rather than creating an entirely new conversation.

To rectify this, we rely on the SCD abstraction 
to remove such surface-level features while maintaining the desired dynamics.
We
prompt the 
LLM
to generate 
a
conversation following the dynamics summarized in the anchor's SCD.
We use human-written (rather than machine-generated) SCDs since they are guaranteed to accurately represent the dynamics as perceived by humans 
\cite{hua_how_2024}, 
while also avoiding circularity with the measures using machine-generated SCDs.
This procedure results in a conversation that, while completely new, 
follows similar dynamics to the anchor conversation, 
forming the anchor-positive pair.

\xhdr{Generating an anchor-negative pair}
To obtain the anchor-negative pair, we must find 
conversations that are known to differ in their dynamics from the anchor.
Drastic differences in outcome can be a good indication that the underlying \dynamics are also different \cite[inter alia]{zhang_conversations_2018, stasi_zooming_2023, dcosta_what_2024, liao_conversational_2023}. 
For each anchor conversation, we pick a different-outcome conversation that is on the same topic and has similar length.
Using
a human-written SCD of the different-outcome conversation, we simulate a conversation that has 
similar dynamics
to it, and thus 
different dynamics
to the anchor.  
We use this simulated conversation to form our anchor-negative pair.\footnote{While in principle we could have directly used the different-outcome conversation to form the  anchor-negative pair, this would have introduced an asymmetry with how the anchor-positive pair is obtained.  One pair would have two real conversations, while the other would have one real and one simulated conversation.
Furthermore, the 
simulation
step becomes important for the sensitivity analysis described below.}

To generate the anchor-positive-negative triplets, we make use of the human-written SCDs provided by 
\citet{hua_how_2024} 
for a subset of 50 ChangeMyView conversations.
These are paired on outcome, such that each conversation that derails into a personal attack is matched with a similar-topic, similar-length conversation that does not.\footnote{According to \citet{hua_how_2024}, the SCDs were written by annotators based on truncated transcripts, such that they could not know the actual outcome of the conversations while writing the summaries.}
This data allows us to create 50 triplets with known relative similarity.

\xhdr{Sensitivity to topical context}
Ideally, a reliable similarity metric for conversational dynamics would not be confounded by topic.
To check which measure best embodies this ideal, we control the topic of the simulated conversations in the triplet to obtain the following conditions:
(1) both the positive and negative conversations are assigned the \textit{same topic} as the anchor; (2) both positive and negative conversations are assigned a \textit{different topic} from the anchor; and (3) an \textit{adversarial} condition in which the positive counterpart has a different topic from the anchor, while the negative counterpart is assigned the same topic.  
The details of the operationalization, including the prompts for identifying and assigning topics are in \autoref{appendix:validation-operationalization}.

\xhdr{Validation results} \autoref{tab:result_table} shows how accurately each similarity measure distinguished between similar (anchor-positive) and dissimilar (anchor-negative) pairs of conversations.
\measurename outperforms all baselines based on matching, in all topic conditions, highlighting the importance of accounting for the order of the interaction patterns through our alignment procedure. 
Furthermore, aligning the SoP to the transcript---and thus allowing for better recall of interaction patterns---results in additional gains over SoP-to-SoP alignment.

Comparing the representations used in each of the matching-based baselines, we see that the SCD representation leads to better accuracy for all measures.
The gains are especially striking in the adversarial topic condition, showing that the abstraction offered by the SCD helps the measures focus on the dynamics and not be distracted by similarities in the topic of the conversation.

\section{Applications}
\label{sec:application}

\begin{figure}[t] 
    \centering
    \includegraphics[width=0.48\textwidth]{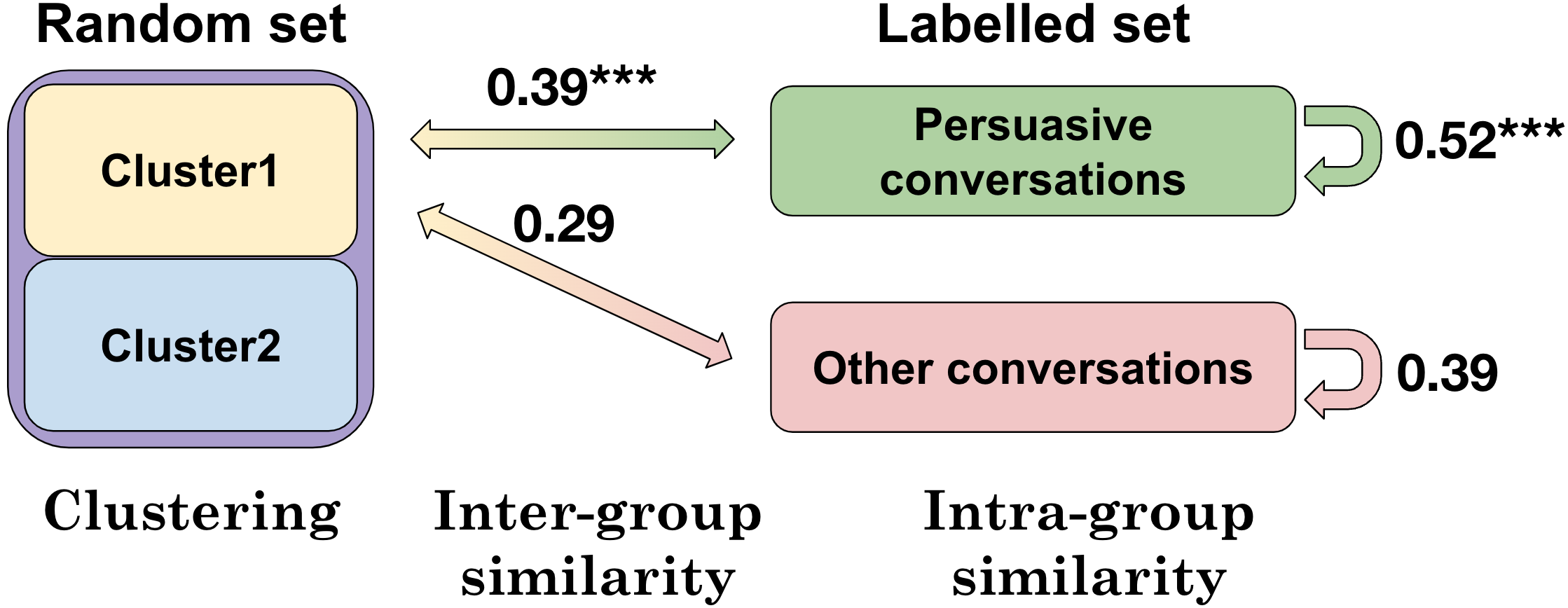}
    \caption{Outline for applying \measurename to different 
    analyses
    supported by similarity measures. Statistically significant differences marked with $^{***}$ ($p<0.001$).} 
    \label{fig:application_diagram}
\end{figure}

\begin{figure}[t] 
    \centering
    \includegraphics[width=0.48\textwidth]{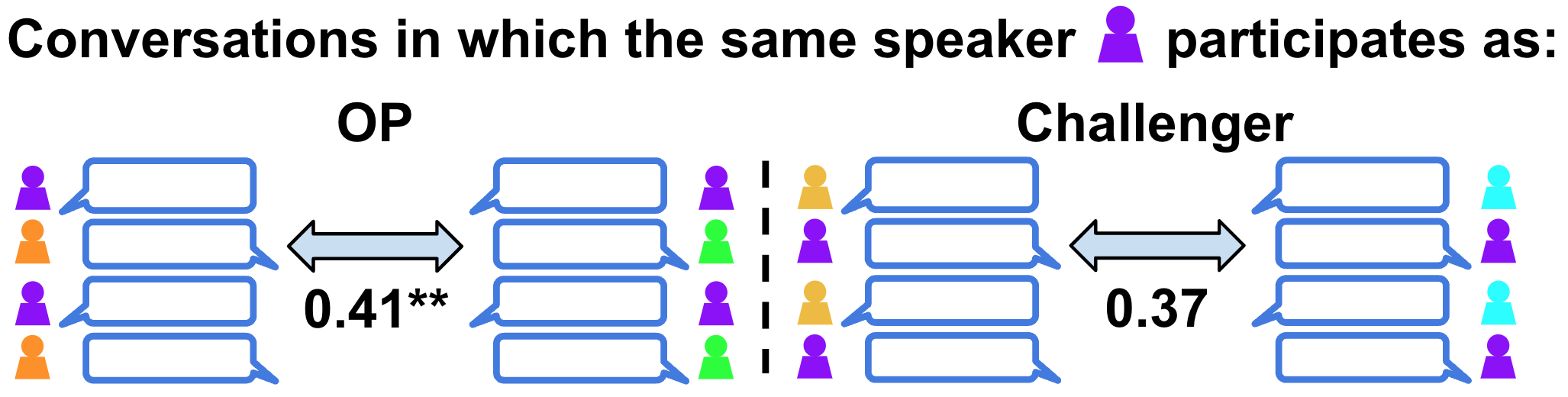}
    \caption{Similarity between two conversations in which a speaker has the role of OP vs. between two conversations in which the \textit{same} speaker has the
    Challenger role.
    The difference is statistically significant ($p < 0.01$).}
    \label{fig:driving_dynamics}
\end{figure}

Having verified its effectiveness, we now demonstrate possible applications of \measurename in analyzing conversational datasets.
We start with showcasing three types of standard data analysis 
techniques
for which a similarity metric is needed---clustering, comparing inter-group similarity, and comparing intra-group diversity---and show that \measurename leads to intuitive results in our online discussions setting (outlined in \autoref{fig:application_diagram}).
We then use our measure to answer new questions about a speaker's tendency to engage in similar dynamics across different conversations and about how a speaker's role in a conversation mediates their influence over its dynamics (\autoref{fig:driving_dynamics}).

\revision{
To ensure that results are not driven by basic structural differences like participant count or conversation length, we focus on conversations that involve only the OP and one Challenger (who always initiates) and that are at least 4 utterances long (and thus are long enough to allow dynamics to develop).
We also consider a stricter length control, in which all conversations are between 4 and 6 utterances.   
Both conditions lead to similar qualitative and numerical results, with small changes in significance levels. 
In what follows we report the results with the strict length control, 
and report those without strict length control in \autoref{appendix:add-application-result}.
}

\subsection{Similarity-based data analysis}

\xhdr{Clustering}
\label{sec:clustering}
To explore common dynamics in CMV,  we cluster a random sample of $200$ conversations from the last year of the data ($2018$) using hierarchical clustering with \measurename.
This involves computing the similarity between all possible pairs of conversations, for a total of $19,900$ comparisons.

We qualitatively characterize the two top-level clusters by exploiting the natural language representation used by \measurename.
Specifically, we aggregate all patterns that receive an alignment score $s_i>0.5$ when measuring the similarity of two conversations in the same cluster.
We compare the aggregated patterns from the two clusters using a Bayesian distinguishing-word analysis \cite{monroe_fightin_2008}, and manually investigate the most distinguishing patterns. 
The results are summarized in \autoref{tab:qualitative_ex_smallest} and examples of corresponding patterns are provided in Tables \ref{tab:qualitative_ex_cluster1} and \ref{tab:qualitative_ex_cluster2} in the Appendix.

The \textit{\textbf{tone}} of the conversation is one of the main components humans consider when describing the conversation's \dynamics \cite{hua_how_2024}.
The tone in \groupone is overwhelmingly positive.
Speakers use negative politeness strategies, such as showing gratitude or confirming the other's points.
They are \textit{collaborative}, building upon each other's argument, and \textit{conciliatory}, apologizing for their misunderstanding or ignorance.
In \grouptwo, on the other hand, the tone is generally characterized by dismissiveness and frustration.
They are \textit{confrontational}---accusing the other speaker of instigating or being passive-aggressive.
In response, the speakers get \textit{defensive} and \textit{sarcastic}---resisting or avoiding direct debates.

\grouptwo's wide range of \textit{\textbf{conversational strategies}} also suggests an argumentative or potentially contentious interaction.
The majority of the speakers express \textit{disagreement} with the other's argument.
The speakers ask a lot of \textit{rhetorical questions} 
in their responses.
They use \textit{straw man fallacies} and \textit{philosophical arguments} and often have to \textit{clarify} their reasoning via \textit{examples} and \textit{analogies}. 
Conversations in \groupone, on the other hand, use detailed \textit{elaboration} to help others understand their arguments.
They are more likely to \textit{agree} and acknowledge the validity of the other speaker's points and concerns; if not, they will \textit{compromise} and concede to points where they share perspectives.

\begin{table}[!t]
    \centering
    \small
    \begin{tabular}{p{1.2cm}p{2.5cm}p{2.5cm}}
        \toprule
        {} & {\textbf{\groupone}} & {\textbf{\grouptwo}}\\
        \midrule
        {Tone} & { negative politeness} & { dismissive}\\
        
        { } & { collaborative } & {sarcastic / defensive}\\
        {} & { conciliatory } & {confrontational}\\
        
\noalign{\vskip 0.3ex}
        {Strategy} & {elaboration } & {straw man fallacy }\\

        {} & { agreement } & {disagreement}\\

        { } & { compromise } & {example / analogy}\\
        { } & {} & { seek clarification }\\
        { } & {} & { philosophical}\\
        { } & {} & { direct responses }\\
\noalign{\vskip 0.3ex}
        { Changes} & { changes in view } & { maintains view }\\

        { } & { lighter tone} & { more contentious}\\
        \bottomrule
    \end{tabular}
    \caption{
    Summary of qualitative analysis of the two identified clusters.
   Examples of corresponding patterns are included in Tables \ref{tab:qualitative_ex_cluster1} and \ref{tab:qualitative_ex_cluster2} in the Appendix.
    }
    \label{tab:qualitative_ex_smallest}
\end{table}

\Dynamics are not only characterized by tone or strategies found in single utterances but also by \textit{\textbf{changes and evolving patterns}}  through multiple utterances.
\groupone's shift in tone is usually toward a 
lighter tone (e.g., serious tone to a humorous one).
Speakers are also more likely to change or revise their claim through discussion.
\grouptwo, on the other hand, becomes more contentious and accusatory, increasing in tension.
Speakers' reluctance to agree often causes initial disagreements to \textit{persist} throughout the conversation, as individuals typically maintain their positions, thereby sustaining the tension.

Overall, this qualitative analysis suggests that the top-level clusters obtained using \measurename correspond to successful and unsuccessful persuasion attempts.
This is expected in an online community focused on debates, further adding face validity to our method.
We can also quantify this distinction by using the labels for successful persuasion ($\Delta$). 
While $\Delta$s are rather rare ($6.5\%$ of conversations in our random sample receive $\Delta$), \groupone and \grouptwo show a significant difference in the proportion of conversations that received a $\Delta$ ($34\%$ vs. $1\%$, 
$p < 0.0001$ 
according to z-test for proportions).

\xhdr{Inter-group similarity}
\label{sec:between_group_sim}
We can further support this interpretation by comparing these automatically detected clusters with a set of conversations that are known to be persuasive. 
We sample a set of $100$ conversations where the OP awarded a $\Delta$ (henceforth \textit{\groupdelta}), and a corresponding set of $100$ corresponding conversations which were not awarded a $\Delta$ (henceforth \textit{\groupnodelta}),
while being triggered by same posts (thus controlling for topic and OP, following \citet{tan_winning_2016}).
There is no overlap between these sets and the random sample used for clustering.

We find that, as suggested by our qualitative analysis, conversations in \groupone are more similar to those that are known to be persuasive (\textit{\groupdelta}) than to those that are not (\textit{\groupnodelta}): mean \measurename of $0.39$ vs. $0.29$, $p<0.001$ per a Mann Whitney U-test. 
It is worth noting that this difference remains significant ($p<0.001$) even if we discard all conversations from \groupone that received a $\Delta$, showing that 
our method can
identify conversations 
that have persuasive-like dynamics even though their persuasiveness is not explicitly acknowledged by the OP. In contrast, \grouptwo's similarity to the two labeled sets is not significantly different.

\xhdr{Intra-group diversity}
\label{sec:intra-group-sim}
Finally, we demonstrate the use of our measure to analyze the diversity of dynamics in a set of conversations by calculating intra-group similarity of \textit{\groupdelta} and \textit{\groupnodelta}, respectively.
Persuasive conversations are significantly more similar to each other than those in which the persuasive attempt fails (mean \measurename $0.52$ vs. $0.39$, $p < 0.001$ according to Mann Whitney U-test; distribution shown in \autoref{fig:within-group-sim}).

\subsection{New investigation: speakers' tendencies}

The dynamics of a conversation 
are the result
of a joint process involving all speakers.
However, 
similar to how
speakers have a tendency to use a specific style \textit{across} different conversations \cite{welch_look_2019,zhang_quantifying_2020}, they may also have a tendency to engage in certain types of dynamics.
In fact, we find evidence to that effect: conversations that share a common speaker have more similar dynamics than those which do not (\measurename $0.37$ vs. $0.35$, $p<0.001$ according to Mann Whitney U-test, comparing similarities within two samples of 1,000 pairs each).

Provided this observation, a natural question arises: in a conversation involving two speakers, whose tendency is more likely to prevail?
In particular, we use our similarity measure to investigate how a speaker's role in a conversation mediates their influence over its dynamics.
Considering the OP and Challenger roles in the CMV setting (Section \ref{sec:datatset}), two hypotheses emerge.
The first is based on situational power \cite{prabhakaran_power_2014}: the OP ultimately decides whether to award a $\Delta$ to the Challenger.
Prior work showed that speakers with higher situational power often influence the other speaker's stylistic \cite{danescu-niculescu-mizil_echoes_2012}, syntactic \cite{boghrati_conversation_2018}, and topical choices \cite{prabhakaran_power_2014}.
Does this influence extend to conversational dynamics?

Alternatively, prior studies emphasize the critical role of persuasion strategies in debates and their outcomes \cite{braca_developing_2023, orazi_verbal_2025}.
The Challenger, by selecting these strategies, may dictate the dynamics.

To distinguish between these two hypotheses, we design a setup that controls for speaker-related confounds, such as demographics that might otherwise spuriously correlate with both their role in the conversation and their influence (\autoref{fig:driving_dynamics}).
We select speakers 
who
participate in at least four conversations, each started by a different post: two in which they take the role of OP and two in which they are 
the Challenger.\footnote{There are $482$ such speakers, considering only dyadic conversations that pass our strict length filters.}

We find that the pair in which the speaker is the OP is more similar than the pair in which they are the Challenger ($0.41$ vs. $0.37, p < 0.01$ according to Wilcoxon signed-rank test).
This suggests that conversation dynamics are more likely to follow the tendencies of the (higher-powered) OP than those of the Challenger, supporting the first hypothesis.\footnote{We note that this result is not a mere consequence of the number of utterances (in both 
conditions, the Challenger speaks slightly more than the OP), or of the success of the persuasion attempt (no statistical difference in $\Delta$'s).}
This result complements the above-mentioned studies 
by providing insights into 
how a speaker's situational role in a conversation mediates their influence on its dynamics.

\section{Application to Other Settings}
\label{sec:application2}
In developing \measurename, we purposely adopted a flexible framework that can accommodate a broad range of conversational dynamics (Section \ref{sec:datatset}).
This design choice facilitates the extension of our measure to new domains with diverse dynamics.

\xhdr{Casual dialogues between friends} To explore this versatility, we apply our method to 50 dialogues from the Friends TV show \cite{chen_character_2016} which are scripted to resemble casual face-to-face dialogues.
The only component of \measurename that needs to be adapted is the prompt for generating the SCDs. 
To tailor these summaries to the dynamics of this particular setting, we adjust the prompt to indicate its casual nature, incorporating a few hand-written examples of summaries (\autoref{fig:scd_prompt_friends} in the Appendix).
We apply the same procedure from Section \ref{sec:application} to provide a qualitative analysis of the two top-level clusters obtained using \measurename in this setting. 
The clusters (\autoref{tab:qualitative_ex_friends} in the Appendix) capture a contrast between (1) more serious conversations, where participants disclose vulnerabilities to seek \textit{validation} or provide \textit{reassurance}, sometimes using \textit{humor} defensively to \textit{deflect} tension, and (2) more \textit{lighthearted} interactions, where \textit{humor} and \textit{inside jokes} are employed to \textit{playfully tease} one another.

\xhdr{Non-English conversations} To examine how our method extends beyond English, we analyze 100 discussions from the German Wikipedia Talk Pages \cite{hua_wikiconv_2018}.
In addition to example hand-written summaries from this dataset, we add to the SCD prompt an instruction to generate the summaries in English.
The resulting clusters are intuitive (\autoref{tab:qualitative_ex_wiki} in the Appendix), distinguishing between (1) conversations in which disagreements are resolved through \textit{polite guidance} supported by \textit{evidence} and \textit{rationales} and (2) conversations where disagreements remain \textit{persistent} and are marked by \textit{sarcasms} and \textit{accusations}.

Overall, this exploration suggests \measurename is versatile enough to be adapted to a variety of conversational settings and to account for different types of conversational dynamics. 
The ability of \measurename to take into account specific aspects of conversational dynamics ultimately depends on the extent to which the SCD-creation procedure can be guided towards surfacing those aspects.

\section{Conclusion}
\label{sec:discussion}
In this work, we introduce a similarity measure for conversational dynamics and develop a validation procedure to compare different representations and alignment methods.
We showcase the measure's utility in the context of an online debate community, adding to the literature on the relation between situational power and influence in conversations.
Our measure joins a growing toolkit of computational methods for conversational analysis. 

In future work, our measure could be used towards a more holistic 
evaluation
of LLMs' conversational ability,
going beyond the quality of each individual reply to compare systems with respect to the dynamics they engender.
\revision{For instance, our metric could be applied to compare the dynamics of mental health therapy conversations held by human therapists versus AI therapists, a timely problem with important societal ramifications.}

Similarity measures like \measurename can also be an important step towards providing conversational-level feedback to AI agents to encourage \textit{dynamics} that are similar to those preferred by humans.
For example,  post-training reinforcement-learning methods could be extended to consider rewards based on similarity with human-preferred dynamics, in addition to human-preferred replies.

\section{Limitations}
\label{sec:limitations}
\measurename should not be regarded as a conclusive measure for conversation dynamic similarity, but as a starting point for better approaches.
It relies on simple prompting for multiple components, and each of them includes non-trivial tasks.
Specifically, we noticed the difficulty of quantifying the alignment of interactional dynamics.
Without a specific rubric, it is difficult to interpret the score that the model outputs.
While our measure provides a short description of the analysis for interpretability, there is much room for future work to systematize the scoring standard and procedure.

Moreover, \measurename requires multiple rounds of generation, which can be very computationally expensive.
The entire transcript of each conversation is used as an input twice to calculate the similarity.
Optimizing the measure would enable it to scale more effectively to larger datasets.

Our proposed validation carries the shortcoming of relying on synthetic data.
Simulated conversations noticeably contained less vulgar and explicit language than real conversations.
Such a difference can lead to a discrepancy in performance when the measure is used with real-life data.
The reliability of the validation process can be improved by enhancing the quality of the simulated conversations.

More broadly, our main analysis is focused on a single domain, which was particularly convenient for developing and validating the measure.
As shown in our exploratory analysis of additional domains, the prompt for generating SCD needs to be adapted in order to apply the method to different settings and languages.
The usefulness of the comparison provided by \measurename hinges on the type of dynamics captures by the SCDs.  
Further work is needed to explore the limits of SCD generation in vastly different conversational settings, and to develop more sophisticated methods that can better capture relevant dynamics.

By releasing our code in a modular fashion, together with demos on several domains, we encourage adaptations and applications to other domains.
A particular challenge is translating our method to multi-modal contexts, exploring how to compare dynamics emerging from vocal (e.g., voice inflection, tone) or visual  (e.g., expressions, gestures) patterns.

Finally, our analysis provides new insights into the role of situational power in conversations.
While in our analysis we control for speaker-specific factors, such as demographics, future work could explore what characteristics beyond the role of the speaker in conversation mediate their influence on the dynamics.
Furthermore, combining our measure with a controlled experiment could complement our observational study to elucidate the causal link between situational power and conversational dynamics.

\textbf{Ethical concerns} associated with LLMs in terms of fairness and bias also extend to \measurename due to its significant dependence on them.
Especially during score assignment, the black-box nature of language models 
is a challenge
without a clear rubric that we can rely on to retrace the logic of the models.
Therefore, \measurename may inadvertently reflect or amplify biases the model was exposed to during training.

\paragraph{Acknowledgments}
We thank the reviewers for their feedback, which in one case resulted in  a conversation with dynamics that are hard to match.
We are grateful for engaging discussions with Team Zissou---including Yash Chatha, Nicholas Chernogor, Tushaar Gangavarapu,  Laerdon Yah-Sung Kim, Lillian Lee, Vivian Nguyen, Luke Tao, Son Tran, and Ethan Xia.
This work was enabled by a Gemma Academic Program GCP Credit Award.
We gratefully acknowledge use of the research computing resources of the Empire AI Consortium, Inc, with support from Empire State Development of the State of New York, the Simons Foundation, and the Secunda Family Foundation.
Cristian Danescu-Niculescu-Mizil was funded in part by the U.S. National Science Foundation under Grant No. IIS-1750615 (CAREER), by Cornell's Center for Social Sciences, by a LinkedIn Research Award, and by a Wikimedia Research Fund Award.
Any opinions, findings, and conclusions in this work are those of the author(s) and do not necessarily reflect the views of Cornell University or the National Science Foundation.

\bibliography{refs}

\appendix
\label{appendix:appendixsection}

\section{Prompts for \measurename}
\label{appendix:ourmeasure_prompts}
In this section, we provide the prompts used in \measurename for all three domains where we applied the measure.
We arrived on these prompts after exploring multiple different versions of the prompt and qualitatively examining the results.

\xhdr{SCD generation}
To generate SCDs for ChangeMyView subreddit conversations, we used the procedural prompt presented in \citet{hua_how_2024}.
The prompt is presented in \autoref{fig:scd_prompt}.
For conversations from German Wikipedia talk-pages, we adapt the prompt to focus on collaboration, as presented in \autoref{fig:scd_prompt_german}.
\textit{Friends} dialogues focus on sentiment and individual characters' intentions, thus the prompt is modified as shown in \autoref{fig:scd_prompt_friends}.

\begin{figure}[t]
\centering
\fbox{
\begin{minipage}{0.45\textwidth}
\raggedright
\small
Write a short summary capturing the trajectory of an online conversation.\\

Do not include specific topics, claims, or arguments from the conversation. The style you should avoid: \\

\textbf{Example Sentence 1:} ``SPK1, who is Asian, defended Asians and pointed out that a study found that whites, Hispanics, and blacks were accepted into universities in that order, with Asians being accepted the least. SPK2 acknowledged that Asians have high household income, but argued that this could be a plausible explanation for the study's findings. SPK1 disagreed and stated that the study did not take wealth into consideration.''\\

This style mentions specific claims and topics, which are not needed.\\

Instead, do include indicators of sentiments (e.g., sarcasm, passive-aggressive, polite, frustration, attack, blame), individual intentions (e.g., agreement, disagreement, persistent-agreement, persistent-disagreement, rebuttal, defense, concession, confusion, clarification, neutral, accusation), and conversational strategies (if any) such as ``rhetorical questions'', ``straw man fallacy'', ``identify fallacies'', and ``appealing to emotions.''\\

The following sentences demonstrate the style you should follow: \\

\textbf{Example Sentence 2:} ``Both speakers have differing opinions and appeared defensive. SPK1 attacks SPK2 by diminishing the importance of his argument and SPK2 blames SPK1 for using profane words. Both speakers accuse each other of being overly judgemental of their personal qualities rather than arguments.''\\

\textbf{Example Sentence 3:} ``The two speakers refuted each other with back and forth accusations. Throughout the conversation, they kept harshly fault-finding with overly critical viewpoints, creating an intense and inefficient discussion.''\\

\textbf{Example Sentence 4:} ``SPK1 attacks SPK2 by questioning the relevance of his premise and SPK2 blames SPK1 for using profane words. Both speakers accuse each other of being overly judgemental of their personal qualities rather than arguments.''\\

Overall, the trajectory summary should capture the key moments where the tension of the conversation notably changes. Here is an example of a complete trajectory summary:\\

\textbf{Trajectory Summary:} \\
Multiple users discuss minimum wage. Four speakers express their different points of view subsequently, building off of each other's arguments. SPK1 disagrees with a specific point from SPK2's argument, triggering SPK2 to contradict SPK1 in response. Then, Speaker3 jumps into the conversation to support SPK1's argument, which leads SPK2 to adamantly defend their argument. SPK2 then quotes a deleted comment, giving an extensive counterargument. The overall tone remains civil.\\

Now, provide the trajectory summary for the following conversation.\\
\textbf{Conversation Transcript:}
\end{minipage}
}
\caption{Procedural prompt for generating SCD \cite{hua_how_2024}}
\label{fig:scd_prompt}
\end{figure}
\begin{figure}[t]
\centering
\fbox{
\begin{minipage}{0.45\textwidth}
\raggedright
\small
Write a short summary capturing the trajectory of a Wikipedia talk-page discussion.\\

Do not include specific article content, titles, policy names, diffs/edits, quotes, or concrete claims. The style you should avoid:\\

\textbf{Example Sentence 1:} ``Speaker1 insisted an article include a particular detail and cited a specific policy by name. Speaker2 countered with a different policy and argued that the section should be removed. Speaker3 referenced a prior version and proposed a precise rewrite.''\\

Instead, do include indicators of sentiments (e.g., sarcasm, politeness, frustration), intentions (e.g., agreement, disagreement, rebuttal, concession, clarification, accusation), and strategies (e.g., consensus attempts, moderation, revert-restore cycles, rhetorical questions, appeals to emotion).\\

The following sentences demonstrate the style you should follow:\\

\textbf{Example Sentence 2:} ``Both speakers hold differing views and become defensive. Speaker1 diminishes the weight of Speaker2’s reasoning, and Speaker2 blames Speaker1 for an uncivil tone. Both accuse each other of focusing on personal traits rather than reasoning.''\\

\textbf{Example Sentence 3:} ``The speakers refute each other with back-and-forth accusations. Persistent fault-finding and critical stances escalate tension and hinder productive discussion.''\\

Overall, the trajectory summary should capture the key moments where the discussion’s tone or coordination changes. Here is an example of a complete trajectory summary.\\

\textbf{Trajectory Summary:}\\
Multiple speakers discuss possible changes. Several present differing stances in sequence, building on and contesting each other’s reasoning. Speaker1 disputes a point from Speaker2, prompting a rebuttal. Speaker3 supports Speaker1, after which Speaker2 defends their position. Later, a speaker references a removed remark and offers an extended counter. Despite friction, the tone remains mostly civil with attempts at consensus.\\

Now, provide the trajectory summary for the following conversation.\\

\textbf{Conversation Transcript:}\\
Now, summarize this conversation. Remember, do not include specific topics, claims, policies, or edits. Instead, capture the speakers’ sentiments, intentions, and strategies. Limit the trajectory summary to 80 words.\\

\textbf{Trajectory Summary (in English):}
\end{minipage}
}
\caption{Procedural prompt for generating SCD on German Wikipedia talk-page discussions}
\label{fig:scd_prompt_german}
\end{figure}

\begin{figure}[t]
\centering
\fbox{
\begin{minipage}{0.45\textwidth}
\raggedright
\small
Write a short summary capturing the trajectory of a casual conversation.\\

Do not include specific topics, events, or arguments from the conversation. The style you should avoid:\\

\textbf{Example Sentence 1:} ``Speaker1 said they had a difficult day at work, and mentioned that their boss was unfair. Speaker2 listened and agreed that bosses can be tough, then suggested they go out for dinner to forget about it..''\\

Instead, do include indicators of sentiments (e.g., warmth, empathy, humor, nostalgia, vulnerability, support), individual intentions (e.g., building rapport, offering reassurance, seeking validation, self-disclosure, active listening, gentle disagreement, creating distance), and conversational strategies (if any) such as ``collaborative storytelling'', ``inside jokes'', ``mirroring emotions'', and ``affectionate teasing''.\\

The following sentences demonstrate the style you should follow:\\

\textbf{Example Sentence 2:} ``Both speakers have similar feelings and appeared mutually supportive. Speaker1 initiates with a moment of self-disclosure, and Speaker2 responds with empathy and validation. Both speakers build on this exchange, strengthening their rapport.''\\

\textbf{Example Sentence 3:} ``The two speakers connected with back-and-forth affectionate teasing. Throughout the conversation, they kept building on each other’s humor with playful remarks, creating a lighthearted and comfortable discussion.''

Overall, the trajectory summary should capture the key moments where the emotional connection of the conversation notably changes. Here is an example of a complete trajectory summary.\\

\textbf{Trajectory Summary:} \\ 
The conversation begins with two speakers exchanging neutral, surface-level comments. Speaker1 then shifts the tone by sharing a personal anecdote, prompting Speaker2 to respond with warmth and empathy. Speaker1 elaborates on their story and their need, but Speaker2 does not extend their support but retracts it.\\

Now, provide the trajectory summary for the following conversation.\\

\textbf{Conversation Transcript:}
Now, summarize this conversation. Remember, do not include specific topics, claims, or arguments from the conversation. Instead, try to capture the speakers' sentiments, intentions, and conversational/persuasive strategies. Limit the trajectory summary to 80 words.\\

\textbf{Trajectory Summary:}
\end{minipage}
}
\caption{Procedural prompt for generating SCD on Friends conversations}
\label{fig:scd_prompt_friends}
\end{figure}
\xhdr{SoP generation}
We prompt a LLM to parse a SCD into a sequence of patterns (SoP).
The prompt is presented in \autoref{fig:sop_parse_prompt}.
With the prompt being highly generic, no adaptation is necessary across conversation domains.

\begin{figure}[t]
\centering
\fbox{
\begin{minipage}{0.45\textwidth}
\raggedright
\small
Here is a trajectory summary of a conversation that lays out how the dynamics of the conversation developed. You need to parse the summary into events in order.\\

Follow the following guidelines:\\
1. Try to maintain the original language of the summary as much as you can.\\
2. Provide your output as a Python dictionary with the following structure:\\
\textit{(Note: Do NOT use markdown, JSON formatting, or code block delimiters.)}\\

\{\\
'0': "" // description of the event
'1': "" \\
...\\
\}

Here is the summary:
\end{minipage}
}
\caption{Prompt for parsing a SCD into sequence of patterns (\sop).}
\label{fig:sop_parse_prompt}
\end{figure}

\xhdr{Score assignment}
The prompt first introduces the general description of the task and describes the format of the input.
The inputs are 1) a dictionary---where the key represents the sequence order of patterns and the value is a description of the pattern identified in SCD---and 2) a transcript of a conversation to compare the dictionary to.
Then, it details the specific instructions the model should follow when assigning similarity score for each pattern. 
Mainly, three main instructions are given:
\begin{enumerate}
    \item The order in which the patterns occur should be highly considered. In other words, we want to reward when the order of the patterns are maintained in the transcript.
    \item Consider whether the transcript closely follow the described sequence. We want to penalize if there are many unrelated patterns or long gaps between the patterns.
    \item The pattern can occur between any speakers, and the specific identities of the speakers do not impact the analysis.
\end{enumerate}

The model is asked to provide a score and a short description of the analysis for each pattern in Python dictionary format.
The prompt can be found at \autoref{fig:score_prompt}.
The prompt can be used across conversation settings.

\begin{figure*}[t]
\centering
\fbox{%
\begin{minipage}{0.9\textwidth}
\raggedright
\small
You will be given a transcript and a list of events describing conversational dynamic and trajectories. You are tasked with determining how closely a predefined sequence of dynamics is seen in a provided conversation transcript, both in occurrence and order.

\textbf{Input}:
- The sequence of events is provided as a dictionary, where:
  - Keys: indicate the order of events, starting from '0'.
  - Values: describe each event.

\textbf{Task:}
- Analysis: Analyze how closely a given transcript follows the sequence of described events. Think and analyze whether you see any part of the transcript resembles the event. Remember that the sequence of events also has to be considered.

- Similarity Score: Give a float score ranging from 0 to 1 based on your assessment of how closely the description of the traject.

  - Order Penalty: If an event occurs before previous events (according to sequence keys), it should be scored significantly lower.
  
  - Proximity of Events: Events in the transcript should closely follow the described sequence. If there are many unrelated events or long gaps between key events, the score should be penalized accordingly.
  
  - Speaker Independence: The event can occur between any speakers, and the actual speaker names do not affect the analysis.
  
\textbf{- Example}:

    - 0: No part of the transcript matches the described event at all.
    
    - 0.35: A part resembles the described event but it occured couple utterances after the previous bullet point event.
    
    - 0.6: A part resembles the described event.
    
    - 1: A part exactly matches the described event explicitly and occurred either at the very first utterance or right after the previous event.

\textbf{Output Format}:
Provide your output as a Python dictionary with the following structure:

(Note: Do NOT use markdown, JSON formatting, or code block delimiters.)

{
    '0': {'analysis': 'ANALYSIS (<=20 words)', 'score': i (0 <= i <= 1) },
    '1': ...
    ...
}
\end{minipage}
}
\caption{Prompt for scoring each pattern in a SoP against a transcript}
\label{fig:score_prompt}
\end{figure*}

\section{Examples of SCD and \sop}
\label{appendix:examples_scd_sop}
\autoref{tab:scd_sop_example} shows three examples of SCD and \sop representation.
They are machine-generated SCD and \sop of real conversations from CMV dataset.
The first two conversations are very similar; the last is very different from the first in terms of dynamics.
The entire transcript of each conversation can be found in \autoref{fig:convo_1_transcript}, \ref{fig:convo_2_transcript}, and \ref{fig:convo_3_transcript}.
\begin{table*}[!ht]
    \centering
    \small
    \begin{tabular}{p{0.5cm}p{6cm}p{7cm}}
    \toprule
    {}&{SCD}&{\sop} \\
    \midrule
    {1}&
    {Speaker2 begins by questioning Speaker1's stance, expressing doubt and using rhetorical questions. Speaker1 clarifies their position, offering an alternative explanation. Speaker2 identifies a perceived inconsistency in Speaker1's statements, suggesting a potential dismissal of authentic experiences and appealing to the importance of further study. Speaker1 reiterates their initial claim with conviction, contrasting two different approaches to evidence and emphasizing a lack of progress in one area.}&
    {
    \begin{enumerate}[nosep]
        \item Speaker2 questions Speaker1 stance, expressing doubt and using rhetorical questions
        \item Speaker1 clarifies their position, offering an alternative explanation
        \item Speaker2 identifies a perceived inconsistency in Speaker1 statements, suggesting a potential dismissal of authentic experiences and appealing to the importance of further study
        \item Speaker1 reiterates their initial claim with conviction, contrasting two different approaches to evidence and emphasizing a lack of progress in one area
    \end{enumerate}
    } \\
    
    {\textcolor{blue}{2}}&
    {\textcolor{blue}{Speaker1 and Speaker2 begin with differing opinions, but maintain a civil tone. Speaker2 attempts to clarify Speaker1's position with a question. Speaker1 responds by elaborating on their stance, providing examples and justifications. Speaker1 aims to clarify their position by providing examples. The conversation remains relatively calm and focused on understanding each other's perspectives.}}
    &{\textcolor{blue}{\begin{enumerate}[nosep]
        \item Speaker1 and Speaker2 begin with differing opinions, but maintain a civil tone
        \item Speaker2 attempts to clarify Speaker1 position with a question
        \item Speaker1 responds by elaborating on their stance, providing examples and justifications
        \item Speaker1 aims to clarify their position by providing examples
        \item The conversation remains relatively calm and focused on understanding each other perspective
    \end{enumerate}}} \\ 
    {\textcolor{red}{3}}&
    {\textcolor{red}{Speaker2 initiates the conversation by recommending a segment. Speaker1 expresses a desire for a concise summary, prompting Speaker2 to claim that a summary would be insufficient. Speaker2 then expresses a negative opinion, using subjective language. Speaker1 responds with agreement and expands on the negative sentiments, while also noting agreement with the underlying message. The overall tone is polite and agreeable.}}
    &{\textcolor{red}{\begin{enumerate}[nosep]
        \item Speaker2 initiates the conversation by recommending a segment
        \item Speaker1 expresses a desire for a concise summary
        \item Speaker2 claims that a summary would be insufficient
        \item Speaker2 expresses a negative opinion, using subjective language
        \item Speaker1 responds with agreement
        \item Speaker1 expands on the negative sentiments
        \item Speaker1 notes agreement with the underlying message
    \end{enumerate}}} \\ 
    \bottomrule
    \end{tabular}
    \caption{
    Example SCD and \sop representations of three conversations. Conversation 2 (colored in \textcolor{blue}{blue}) is a similar conversation (positive) of Conversation 1 (\measurename assign score of 0.544). Conversation 3 (colored in \textcolor{red}{red}) is a non-similar conversation (negative) of Conversation 1 (score of 0.112). 
    See \autoref{fig:convo_1_transcript}, \ref{fig:convo_2_transcript}, and \ref{fig:convo_3_transcript} for the entire transcript of each conversation. 
    }
    \label{tab:scd_sop_example}
\end{table*}

\begin{figure*}[htbp]
\centering
\fbox{%
\begin{minipage}{0.9\textwidth}
\raggedright
\small
\textbf{SPEAKER1}: There's really no easier way of putting it. Can you really expect me to believe people that have these instances where "oh my friend and I..." or "oh I saw it but umm nobody else what there!" C'mon now. Seriously? Why would you believe anything without evidence? 

Why? Like...why. I just don't get it. I'm not sure I understand the reasoning behind trying to scare other people and stuff. And those who get spooked are just as lame. I slept in 2 "haunted" houses by myself just to prove a point (and also for money from a bet!) and nothing happened. And yes, I recorded the whole thing with a GoPro. I went to sleep, nothing happened. Nothing strange has ever happened to me and I've been to numerous places where there's been "reported sightings!!!" ( o0o0o0o0o0o0o0 so scary).

I'm just sick of all these people claiming this stuff for attention or letting their minds play tricks on them. I bet all of them haven't even gotten enough sleep either.

EDIT: Look what I found! [hyperlink]

> *This is a footnote from the CMV moderators. We'd like to remind you of a couple of things. Firstly, please* ***[read through our rules]([hyperlink])***. *If you see a comment that has broken one, it is more effective to report it than downvote it. Speaking of which,* ***[downvotes don't change views]([hyperlink])****! Any questions or concerns? Feel free to* ***[message us]([hyperlink]. *Happy CMVing!*

\textbf{SPEAKER2}: You're basically arguing that the people who claim to have seen something when nobody else is around was lying. Right? Or do you think that there is simply a logical explanation for what they claim to have seen? I doubt people would just outright lie about an experience like that.

Near-Death Experiences are a great example. We know they are definitely real now, but had we carried this attitude of "why should we believe this happened to you", then we would have missed out on an incredibly fascinating field of scientific study.

\textbf{SPEAKER1}: I think there's a logical explanation, and that it doesn't involve "ghosts" and is more inline with our brain chemistry and such. Essentially different parts of our brain doing varying things, be they in error or not. 

Look at sleep paralysis. That's cause for waking hallucinations - and we understand it and can explain it.

\textbf{SPEAKER2}: Ok but this sentence from your OP

>;why would you believe anything without evidence?

is promoting a very different belief from what you just said here. The quoted statement makes it sound like you think these people made it all up. If their brain chemistry made them see something, then that IS an authentic experience; we just called it something inaccurate.

This is important because if we think it's just all baloney, we would never study it further.

>Look at sleep paralysis. That's cause for waking hallucinations - and we understand it and can explain it.

And we never would have figured this out if we approached a statement like "I saw you walk across the room while I was sleeping" with a statement like "why would we believe anything like this without evidence?"

\textbf{SPEAKER1}: Allow me to reiterate: Ghosts do not exist. 

While some have tried to prove they have, nothing happened to our benefit. 

We also have discovered much else because of the same approach. It has been tried to be proven, but failed. Sleep paralysis was a confirmed thing that we kept looking into, just like ghosts. Except one is now much further because of the thread of evidence that we had in the first place. The other doesn’t.
\end{minipage}
}
\caption{Full transcript of Example Conversation 1}
\label{fig:convo_1_transcript}
\end{figure*}

\begin{figure*}[htbp]
\centering
\fbox{%
\begin{minipage}{0.9\textwidth}
\raggedright
\small
\textbf{SPEAKER1}: To say Player Unknown's Battle Grounds (PUBG) and Fortnite Battle Royal (Fortnite) have gotten huge is a vast understatement. PUBG first dominated the scene earlier this year by being a definitive addition to the genre, then Fortnite stole the limelight by addressing  the problems (mainly developer integrity and system performance) PUBG had. Both are still going strong with their own audiences, art styles, design choices, and most importantly, eSports leagues. Big name teams like Cloud9 and Natus Vincere are hopping on board PUBG's league, and Fortnite's publisher, Epic Games, announced their 100 million contribution to prize pools for competetions for the next year.

I think it's all bullshit.

Any game in the battle royal genre is inherently unbalanced.  RNG luck is too big of a factor in these games, making every game unfair regardless of the circumstances.  Where you can drop at the beginning of the match, who's sitting next to you on the plane, what guns will be on the ground waiting for you, and when/where the supply drops are, are all random.  Success in the game is determined more by luck than skill  there's nothing that even best player can do when they finally land only to be blasted in the face by someone else with the shotgun that just so happened to be closer to them.

This brings me to my other point on it's effect on the eSports scene.  The games that have defined eSports CounterStrike, DoTa 2, League of Legends, etc.  draw many parallels to physical sports.  They require skills that can be practiced, and can benefit from strategies, techniques, and teamwork, similar to a real sport.  I have a phrase that I've been waiting to say to someone that says otherwise: "This isn't competitive Candy Crush."  I've argued against people that try to overgeneralize video games as "sitting on their ass hitting buttons," overlooking the mechanical skills and knowledge of the game required to do well.  I fear that if PUBG and Fortnite takes off in a competitive sense, the amount of luck present in the game will undermine the games I listed earlier those built from the ground up to give players a level playing field  as being easier than they are.

eSports is a well\-established industry at this point, and to say it's here to stay should be a given.  But with the notion of the BR genre making it's presence known, I do have my concerns on how people think about eSports as a whole.

Edit:  I should probably clarify, my point on RNG in the BR genre is that RNG is *too far embedded* into the games to make it competitive, and not enough of it can be mitigated to make things a fair fight. RNG is fine in other games, so long as they can be mitigated.  

I should also clarify that when I say RNG, I mean a true Random Number Generator. Variances from other sources I have no problem with.

\textbf{SPEAKER2}: PUBG is definitely not eSports ready but not because of the core RNG mechanic of the game. Every sport or game has factors outside of the player's control, and part of being a good competitor is being able to prepare and react to it. As long as the RNG component can't be hacked or manipulated then it is fair by definition. It may be merely that the structure of the competition needs to take into consideration the RNG element, for example a PUBG tournament should be based on several matches and not just single elimination so that player skill has a chance to shine. 

\textbf{SPEAKER1}: I understand there are factors outside of the player's control in any activity.  Having some of these factors being decided by a computer is what I'm against.  If some of these factors can at least be mitigated (eg, rain at an event can be fixed by a stadium with a roof, weapon spread can be disabled server-side), then I'm okay with it, but shooting someone's head and missing because a computer decided I don't get to kill someone today is infuriating, and in my opinion not fun to watch.

\textbf{SPEAKER2}:
> but shooting someone's head and missing because a computer decided I don't get to kill 

So do you have a problem with Battle Royal games or just with the gun mechanics? Overwatch has RNG bullet spread too though obviously more consistent. 

\textbf{SPEAKER1}: I have a problem with the RNG that's in both.  The RNG that determines gun inaccuracy, as well as the RNG that determines which plane you're on and what/where weapons/boxes will spawn.

I'm a bit rusty on Overwatch, the only hitscans I can think of that have spread would be Soldier 76, Tracer, McCree's "Fan the Hammer," Roadhog, and Reaper.

Tracer, Roadhog, and Reaper, and McCree's FtH are meant to be used up close, where RNG doesn't matter.  McCree's basic attack is 100 amp;37; accurate with a slow rate of fire, which brings me to Soldier 76.  His bloom can be worked around by simply bursting/tapping his rifle, which I'm fine with.

In games with weapon inaccuracy, what makes a player skilled is his ability to circumvent/mitigate the inaccuracy.  In CS:GO, where moving makes your gun shoot everywhere on your screen, movement comes with several options to mitigate movement inaccuracy (like counter strafing).
\end{minipage}
}
\caption{Full transcript of Example Conversation 2}
\label{fig:convo_2_transcript}
\end{figure*}

\begin{figure*}[htbp]
\centering
\fbox{%
\begin{minipage}{0.9\textwidth}
\raggedright
\small
hyperlink: I've been a member for a year, ever since I began educating myself about firearms, took extensive training, and bought three. I've now also passed enhanced background checks and earned concealed carry permits in three states.

I haven't seen any news items with good arguments against the NRA that hold up on scrutiny. Every article I see is,  "ignore what they're saying; here's what they really mean". You can imagine how that's unconvincing.

Plus, the latest CNN (?) town square with students, Dana Loesch and politicians was the worst of mob theater. Nothing there for me but confirmation in my beliefs.

As an organization for its members, I like everything the NRA does: they change with the times [sponsoring great vloggers like Colion Noir]([hyperlink]), offering insurance and legal help, and supporting victims of gov't gun confiscation. [Example video]([hyperlink]), [case info]([hyperlink]).

About me: I'm a member of both the NRA and PETA. I'm politically moderate, After decades believing the "conventional wisdom" about these and other groups, I started deep diving into the supporting facts behind the frequent hit pieces about them. And I found that most (all?) fall apart under scrutiny.

> *This is a footnote from the CMV moderators. We'd like to remind you of a couple of things. Firstly, please* ***[read through our rules]([hyperlink])***. *If you see a comment that has broken one, it is more effective to report it than downvote it. Speaking of which,* ***[downvotes don't change views]([hyperlink])****! Any questions or concerns? Feel free to* ***[message us]([hyperlink])***. *Happy CMVing!*

\textbf{SPEAKER2}: Watch the enost recent segment John Oliver did on them. It's pretty interesting. 

SPEAKER1: Thanks - could you sum it up in a sentence or two?

\textbf{SPEAKER2}: Not well enough. It's like twenty minutes. One good thing to notice about them though is that they are no different than an infomercial channel. They profit off of their beliefs which is why their ads-IMO- are so cringey with their intenseness. I stopped following gun channels on YouTube who ran their ads. Don't regret it. 

\textbf{SPEAKER1}: I'm also not a fan of their videos (or any) with the threatening soundtrack, etc. etc. Also the excessive branding and intro screens. Yeah, I avoid those. I pretty much always agree with the message, though.

\end{minipage}
}
\caption{Full transcript of Example Conversation 3}
\label{fig:convo_3_transcript}
\end{figure*}

\section{Validation Details}
\label{appendix:validation-operationalization}
\subsection{Baseline Implementation Details}
\xhdr{Cosine similarity} We use a pre-trained sentence BERT model `all-MiniLM-L6-v2’ ($22.7$M parameter in size) to map either the entire transcript or the generated SCD of the conversation into a 384 dimensional dense vector space. 
Text that is longer than 256 tokens are truncated.
The similarity between two conversations is measured by calculating the cosine similarity of their embeddings.

\xhdr{BERTScore} We use a distilled version of the BERT base model ($67$M parameters) \cite{sanh_distilbert_2020} and Huggingface's BERTScore pipeline to calculate the similarity score. 

\xhdr{Naive prompting} We use `chatgpt-4o-latest' model via OpenAI API. 
The prompt includes the definition of conversation trajectory, specific instructions to consider, output format.
The prompt for comparing transcripts can be found at \autoref{fig:naive_prompt}.
The prompt for comparing SCDs can be found at \autoref{fig:naive_prompt_scd}.

\begin{figure*}[t]
\centering
\fbox{%
\begin{minipage}{0.9\textwidth}
\raggedright
\small
Compare the following two online conversations and rate their similarity on a scale from 1 to 100, based on their trajectory.\\

\textbf{Definition of Trajectory}\\
The trajectory of a conversation refers to its dynamics, including:\\
-- \textbf{Changes in tone} (e.g., neutral to argumentative, formal to casual, sarcastic or sincere).\\
-- \textbf{Patterns of interaction} (e.g., back-and-forth exchanges, long monologues, interruptions).\\
-- \textbf{Conversation strategies} (e.g., persuasion, questioning, storytelling).\\
-- \textbf{Order of the above trajectory events}\\

\textbf{Ignore}:\\
-- The topics discussed.\\
-- Specific factual content.\\

\textbf{Output Requirements}\\
Return a JSON object containing:\\
-- \texttt{"sim\_score"} (int): A similarity score between 1--100, representing how similar the conversations are in \textbf{trajectory}.\\
-- \texttt{"reason"} (string, \textless=30 words): A brief explanation of why the score was given, referencing key conversational dynamics.\\

\textbf{Output Format (JSON)}\\

\{\\
    sim\_score: int,\\
    reason: brief explanation under 30 words\\
\}

\textbf{Conversations}\\
\textbf{Conversation 1:} \\
\textbf{Conversation 2:}
\end{minipage}
}
\caption{Prompt for naive prompting baseline}
\label{fig:naive_prompt}
\end{figure*}

\begin{figure*}[t]
\centering
\fbox{%
\begin{minipage}{0.9\textwidth}
\raggedright
\small
Compare the following two summary of conversation dynamics (SCD) of two online conversations, rate the similarity of the two conversations on a scale from 1 to 100, based on their persuasion trajectory reflected in the SCDs.\\

\textbf{Definition of Trajectory}\\
The trajectory of a conversation refers to its dynamics, including:\\
-- \textbf{Changes in tone} (e.g., neutral to argumentative, formal to casual, sarcastic or sincere).\\
-- \textbf{Patterns of interaction} (e.g., back-and-forth exchanges, long monologues, interruptions).\\
-- \textbf{Conversation strategies} (e.g., persuasion, questioning, storytelling).\\
-- \textbf{Order of the above trajectory events}\\

\textbf{Ignore}:\\
-- The topics discussed.\\
-- Specific factual content.\\

\textbf{Output Requirements}\\
Return a JSON object containing:\\
-- \texttt{"sim\_score"} (int): A similarity score between 1--100, representing how similar the conversations are in \textbf{trajectory} based on the SCDs.\\
-- \texttt{"reason"} (string, \textless=30 words): A brief explanation of why the score was given, referencing key conversational dynamics.\\

\textbf{Output Format (JSON)}\\

\{\\
    sim\_score: int,\\
    reason: brief explanation under 30 words\\
\}

\textbf{Conversations}\\
\textbf{Conversation 1 SCD:} \\
\textbf{Conversation 2 SCD:}
\end{minipage}
}
\caption{Prompt for naive prompting baseline with SCDs}
\label{fig:naive_prompt_scd}
\end{figure*}

\subsection{Simulating conversation}
\label{appendix:simulating_conversation}
For simulating conversations, we use a snapshot of OpenAI’s GPT-4o-mini model from July 18th, 2024 \cite{achiam_gpt-4_2024}, accessed via the OpenAI API due to its cost-efficiency.
We ask a language model to recreate an online conversation, given the topic of the conversation and a summary of it's dynamics.
The prompt used for conversation simulation is included in \autoref{fig:simulation_prompt}.

\xhdr{Alternative approach to simulating}
Initially, we simulated conversations with similar \dynamics by inputting a conversation transcript.
The prompt is included in \autoref{fig:sim_w_transcript_prompt}.
The simulated conversations, however, were trivially similar to each other in how they carried out the \dynamics.
They would often copy the exact same sentence structure, sometimes repeat the same words or phrases used in the original transcript, or speaker order---even when it was instructed to generate a conversation with a different topic.
Here are some examples:\\
Example 1: 
\begin{itemize}[itemsep=0pt, leftmargin=*]
    \item Original transcript: ``Even if that's true in the election, it changes the overall vote split between the parties.''
    \item Simulation: ``Even if that’s true, the market is shifting.''
\end{itemize}
Example 2:
\begin{itemize}[itemsep=0pt, leftmargin=*]
    \item Original transcript: ``isnt that what the king of England wanted from the colonies when we rebelled?''
    \item Simulation: ``Isn’t that kind of like buying a car that’s cheaper upfront but costs more in gas and repairs?''
\end{itemize}

Such observations highlighted the need for a simulation method that provides the model with the dynamics it needs to follow while not exposing it from the original transcript. 
\begin{figure}[t]
\centering
\fbox{
\begin{minipage}{0.45\textwidth}
\raggedright
\small
You are given a task to recreate an online conversation that occurred on reddit. Here is a list of information you are given.\\
1. Topic of the conversation: \{topic\}\\
2. The original conversation that which the conversation trajectory you should follow: \{transcript\}\\

\#\#\# \textbf{Definition of Trajectory}
The trajectory of a conversation refers to its **dynamics**, including:
- **Changes in tone** (e.g., neutral to argumentative, formal to casual, sarcastic or sincere).\\
- **Patterns of interaction** (e.g., back-and-forth exchanges, long monologues, interruptions).\\
- **Conversation strategies** (e.g., persuasion, questioning, storytelling).\\
- **Order of the above trajectory events**\\

\#\#\# \textbf{Ignore}:
- The topics discussed.\\
- Specific factual content.\\

In your recreated conversation, each utterance of the transcript should be formatted as the following:
Speaker\_ID (e.g. "SPK2") :\\

\#\textbf{Output}
Add your recreated conversation. Only generate the transcript of the conversation. 
\end{minipage}
}
\caption{Prompt for simulating conversation with transcript}
\label{fig:sim_w_transcript_prompt}
\end{figure}

\begin{figure}[t]
\centering
\fbox{
\begin{minipage}{0.45\textwidth}
\raggedright
\small
You are given a task to recreate an online conversation that occurred on reddit. Here is a list of information you are given.

1. Topic of the conversation: \{topic\}\\
2. Trajectory summary that summarizes the conversational and speakers' dynamics: \\
\{trajectory\_summary\}

Each utterance of the transcript should be formatted as the following:\\
Speaker\_ID (e.g. "SPK2") : Add text of the utterance\\

\#Output
Add your recreated conversation. Only generate the transcript of the conversation. 
\end{minipage}
}
\caption{Prompt for simulating conversation}
\label{fig:simulation_prompt}
\end{figure}

\subsection{Topic setting during simulation}
\label{appendix:topic_setting}
\begin{figure}[t]
\centering
\fbox{
\begin{minipage}{0.45\textwidth}
\raggedright
\small
Here are two conversations of the same topic. Summarize the topic of the conversations in a concise phrase that accurately captures the main subject being discussed.\\
Here is the transcript of the first conversation:\\
\{transcript1\}\\

Here is the transcript of the second conversation:\\
\{transcript2\}\\

Now, write the topic of the conversation in a concise phrase:\\
\end{minipage}
}
\caption{Prompt for identifying the topic of the conversation.}
\label{fig:topic_identify_prompt}
\end{figure}

We first need to identify the topic of the anchor conversation to run a topic sensitivity analysis.
We prompt a model to identify the topic of a conversation-pair, as we have a paired dataset \cite{hua_how_2024}, whose pairs have the same topic.
The prompt is provided in \autoref{fig:topic_identify_prompt}.

For the \textit{same topic} setting, the topic identified for an anchor conversation is used as the imposed topic to simulate its positive and negative counterparts.
For the \textit{different topic} setting, topics identified from the $50$ conversations in the dataset are first shuffled. 
Each anchor is then assigned one of these shuffled topics, which serves as the specified topic for simulating its positive and negative counterparts, while ensuring that the shuffle topic is different from the anchor's original identified topic.
For the \textit{adversarial} setting, a topic obtained through the shuffling process described in the \textit{different topic} setting is used for simulating the positive counterpart, while the anchor's original identified topic is used for simulating the negative counterpart.

\section{Additional Validation Results}
\label{appendix:additional_val}

We validated our measure using OpenAI's gpt-4o model \cite{achiam_gpt-4_2024} as well.
We also ran all baseline using gpt-4o generated SCDs.
The result is summarized in \autoref{tab:additional_result}.

\begin{table*}[t]
    \centering
    \begin{tabular}{p{2.3cm}|p{1.8cm}p{1cm}p{1.8cm}p{1.8cm}p{3cm}}
    \toprule
    {  Measure} & \multicolumn{2}{c}{  \measurename} & {  cosine sim.} & {  BERTScore} & {  Naive Prompting}\\
    {Representation} & {SoP+Trx} & {\sop} & {SCD} & {SCD} & {SCD} \\
    \midrule
    {  same topic} & {  \textbf{92}\%} & {82\%} & {  72\%} & {  70\%}  & {  70\%}  \\
    { different topic} & {  \textbf{98}\%} & { 74\% } & {  76\%}  & {  66\%}  & {  64\%} \\
    {  adversarial} & {  \textbf{96}\%} & {72\%} & {  64\%}  & {  70\%}  & {  60\%} \\
    \bottomrule
    \end{tabular}
    \caption{\label{citation-guide} Accuracy of each measure in our validation setup using gpt-4o. The accuracy of baseline is when using gpt-4o's generated SCD as its input.}
    \label{tab:additional_result}
\end{table*}

\section{Qualitative Examples}
\label{appendix:qualitative_examples}

\autoref{tab:qualitative_ex_cluster1} and \autoref{tab:qualitative_ex_cluster2} includes multiple examples of identified patterns in each cluster from the CMV conversations selected with strict length control as reported in the main paper. \autoref{tab:qualitative_ex_cluster1_less_restrict} and \autoref{tab:qualitative_ex_cluster2_less_restrict} includes examples of patterns from each cluster of CMV conversations selected with less strict length control, as discussed in \autoref{appendix:add-application-result}. \autoref{tab:qualitative_ex_friends} includes examples and identified patterns in each cluster from the \textit{Friends} dialogues and \autoref{tab:qualitative_ex_wiki} includes those from the German Wikipedia talk-pages conversations.
\begin{table*}[!ht]
    \centering
    \small
    \begin{tabular}{p{0.2cm}p{1.3cm}p{5cm}p{7cm}}
        \toprule
        {  \#} & {  Category} & {  Dynamics} & {  Examples}\\
        \midrule
        {  1} & {  Tone} & {  \textbf{negative politeness}} & {  SPK1 expresses gratitude for the validating response.}\\
        {  } & {  } & {  (gratitude, thanks, appreciation)} & {  SPK1 expresses empathy and appreciation for SPK2 insight.}\\
        
        {  } & {  } & {  \textbf{collaborative}} & {  SPK1 and SPK2 build upon each other point.}\\
        {  } & {  } & {  (collaborative, build upon)} & {  The conversation maintains a collaborative sentiment throughout.}\\

        {  } & {  } & {  \textbf{conciliatory}} & {  SPK1 acknowledges new information.}\\
        {  } & {  } & {  (acknowledgement, acknowledges, apologizing)} & {  SPK1 apologizes for misunderstanding and offers a polite suggestion for future communication}\\
\noalign{\vskip 0.6ex}

        {  } & {  Strategy} & {  \textbf{elaboration}} & {  SPK2 introduces information}\\
        {  } & {  } & {  (specific, detailed, information, informative)} & {  SPK2 begins by providing a detailed and informative response, seemingly intending to persuade SPK1.}\\

        {  } & {  } & {  \textbf{agreement}} & {  SPK1 expresses agreement and appreciation.}\\
        {  } & {  } & {  (agrees, agreement, validate)} & {  SPK2 attempts to validate SPK1 concerns.}\\

        {  } & {  } & {  \textbf{compromise}} & {  SPK2 offers a revised premise.}\\
        {  } & {  } & {  (compromise,  concedes, concession)} & {  SPK1 initially agrees with SPK2 point but expresses a reservation, seeking a compromise.}\\

\noalign{\vskip 0.6ex}

        {  } & {  Changes} & {  \textbf{changes in perspective}} & {  SPK2 offers a revised premise.}\\
        {  } & {  } & {  (revised, change)} & {  SPK1 then conceded, acknowledging the validity of SPK2 point and expressing a change in perspective.}\\

        {  } & {  } & {  \textbf{shift to lighter tone}} & {  SPK2 shifts to a more agreeable tone.}\\
        {  } & {  } & {  } & {  SPK1 shifts the tone from serious concern to a more humorous outlook.}\\
        \bottomrule
    \end{tabular}
    \caption{
    Qualitative analysis of dynamics of Cluster 1 from ChangeMyView conversations (with strict length control). 
    Phrases in parentheses are distinguishing words used during the analysis.
    }
    \label{tab:qualitative_ex_cluster1}
\end{table*}
\begin{table*}[!ht]
    \centering
    \small
    \begin{tabular}{p{0.2cm}p{1.3cm}p{5cm}p{7cm}}
        \toprule
        {  \#} & {  Category} & {  Dynamics} & {  Examples}\\
        \midrule
        {  2} & {  Tone} & {  \textbf{dismissive}} & {  SPK2 begins by disagreeing...
        using a dismissive tone.}\\
        {  } & {  } & {  (frustrated, dismissive)} & {  SPK2, maintaining a dismissive and sarcastic tone, expresses persistent disagreement.}\\
        
        {  } & {  } & {  \textbf{sarcastic}} & {  SPK2 begins with a rhetorical question, seemingly sarcastic.}\\
        {  } & {  } & {  (sarcasm, sarcastically)} & {  SPK2 responds with sarcasm and attempts to clarify the definition of a term used by SPK1.}\\

        {  } & {  } & {  \textbf{defensive}} & {  SPK1 expresses defensiveness.}\\
        {  } & {  } & {  (defensive, resists)} & {  SPK1 responds defensively, limiting the scope of the discussion and questioning SPK2 reasoning.}\\

        {  } & {  } & {  \textbf{confrontational}} & {  SPK1 maintains a confrontational stance.}\\
        {  } & {  } & {  (accuses, 
        blame, confrontational)} & {  SPK2 accuses SPK1 of using a straw man fallacy.}\\

\noalign{\vskip 0.6ex}
        {  } & {  Strategy} & {  \textbf{straw man fallacy}} & {  SPK2 uses a sarcastic tone and straw man fallacy.}\\
        {  } & {  } & {  (straw man)} & {  SPK1 then uses a straw man fallacy, misrepresenting SPK2 argument to attack it.}\\

        {  } & {  } & {  \textbf{philosophical argument}} & {  SPK1 responds with a philosophical argument.
        }\\
        {  } & {  } & {  (philosophical argument/concept/difference)} & {  SPK2 defends their position, identifying what they believe is a core philosophical difference with SPK1.}\\

        {  } & {  } & {  \textbf{providing examples}} & {  SPK2 attempts to clarify their position using examples.
        }\\
        {  } & {  } & {  (examples, example)} & {  SPK1 continues to disagree, providing counter-examples and expressing skepticism.}\\

        {  } & {  } & {  \textbf{analogy}} & {  SPK2 initiates the conversation with a hypothetical scenario.
        }\\
        {  } & {  } & {  (analogy, analogies, hypothetical)} & {  SPK1 accuses SPK2 of not taking the conversation seriously, while also clarifying their stance.}\\

        {  } & {  } & {  \textbf{seeking clarification}} & {  SPK2 initially expresses confusion and seeks clarification.
        }\\
        {  } & {  } & {  (confusion, lack of understanding, seeking clarification)} & {  SPK1 expresses confusion and disagreement with SPK2 premise.}\\

        {  } & {  } & {  \textbf{disagreement}} & {  SPK2 quickly introduces a contrasting viewpoint.
        }\\
        {  } & {  } & {  (disagrees, disagreement, contrasting)} & {  SPK1 immediately expresses disagreement with the definition.}\\

        {  } & {  } & {  \textbf{direct responses}} & {  SPK1 immediately disagrees, using statistics to justify 
        .
        }\\
        {  } & {  } & {  (direct, directly, immediately, quickly)} & {  SPK2 directly disagrees with SPK1, asserting a factual error and expressing shock.}\\
\noalign{\vskip 0.6ex}
        {  } & {  Changes} & {  \textbf{maintains perspective}} & {  SPK1 maintains a 
        negative tone towards specific actors.
        }\\
        {  } & {  } & {  (continues, maintains strong negative, persists)} & {  SPK1 continues to disagree, using another analogy to defend their position.}\\

        {  } & { } & {  \textbf{shift to contentious tone}} & {  SPK1 shifts from concession to disagreement.
        }\\
        {  } & {  } & {  } & {  SPK1 shifts to a more accusatory tone, implying a lack of justification.}\\
        \bottomrule
    \end{tabular}
    \caption{
    Qualitative analysis of dynamics of Cluster 2 from ChangeMyView conversations (with strict length control). 
    Phrases in parentheses are distinguishing words used during the analysis.
    }
    \label{tab:qualitative_ex_cluster2}
\end{table*}

\section{Additional Applications Results}
\label{appendix:add-application-result}
\xhdr{Application result without strict length control}
We report here the experiment results equivalent to those presented in Section~\ref{sec:application} without the strict length control (maintaining all the other filters).  

\begin{figure}[t] 
    \centering
    \includegraphics[width=0.48\textwidth]{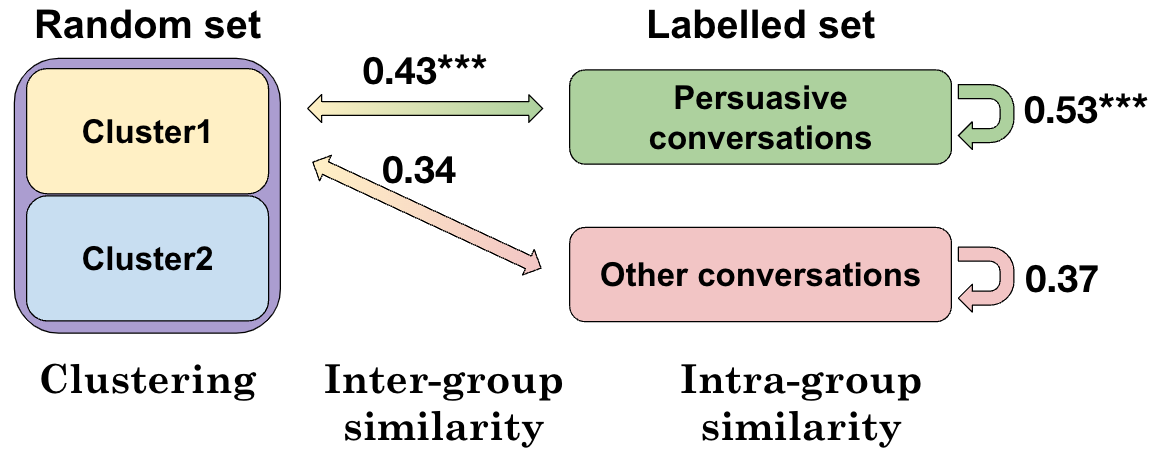}
    \caption{Outline for applying \measurename to different 
    analyses
    supported by similarity measures with less strict length control. Statistical significant differences marked with $^{***}$ ($p<0.001$).} 
    \label{fig:application_diagram_addition}
\end{figure}

We conduct clustering and then compare inter-group similarity as well as intra-group diversity (outlined in \autoref{fig:application_diagram_addition}).
We first cluster a random sample of 200 conversations from the year 2018, using hierarchical clustering with ConDynS.
The qualitative analysis of the two top-level clusters, focusing on aggregated patterns with alignment scores $s_i > 0.5$, reveals a similar separation in the use of tone, conversational strategies, and evolving dynamics compared to the results obtained under stricter length control.
Illustrative examples of the corresponding patterns are provided in \autoref{tab:qualitative_ex_cluster1_less_restrict} and \autoref{tab:qualitative_ex_cluster2_less_restrict}.
We also observe distinctions between the two top-level clusters when considering persuasion success labels ($\Delta$).
In this random set, 13\% of conversations are labeled with $\Delta$.
Cluster 1 and Cluster 2 differ significantly in proportion of conversations receiving $\Delta$ (35\% vs. 3\%, $p < 0.0001$ according to z-test for proportions).
We further sample 100 conversations labeled with $\Delta$ (\textit{set $\Delta$}) and 100 without ($\textit{set $\neg \Delta$}$). 
We find that conversations in Cluster 1 are more similar to \textit{set $\Delta$} than to \textit{set $\neg \Delta$} (mean ConDynS 0.43 vs. 0.34, $p < 0.001$, Mann–Whitney U-test).
Moreover, \textit{set $\Delta$} conversations are significantly more similar to each other than \textit{set $\neg \Delta$} conversations (mean ConDynS 0.53 vs. 0.37, $p < 0.001$, Mann–Whitney U-test).

\begin{figure}[t] 
    \centering
    \includegraphics[width=0.48\textwidth]{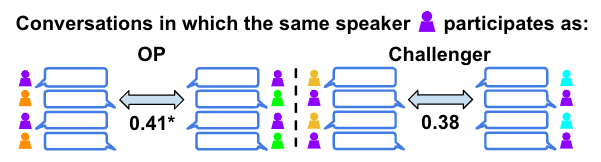}
    \caption{Similarity between two conversations in which a speaker has the role of OP vs. between two conversations in which the \textit{same} speaker is the 
    Challenger.
    The difference is statistically significant ($p < 0.05$). Conversations are filtered with less strict length control.}
    \label{fig:driving_dynamics_addition}
\end{figure}

To investigate speakers' tendencies, we compare the similarity between 1,000 pairs of conversations sharing a common speaker with the similarity between 1,000 pairs without a shared speaker, finding that that pairs sharing a common speaker demonstrate more similar dynamics than those without a shared speaker (ConDynS 0.39 vs. 0.37, $p < 0.01$, Mann–Whitney U-test).
We then focus on the 486 speakers who each participated in at least four conversations---two as the OP and two as the Challenger---and we further observe that pairs where the speaker serves as OP are more similar than pairs where the same speaker serves as Challenger (0.41 vs. 0.38, $p < 0.05$ according to a Wilcoxon signed-rank test).
The setup described in \autoref{fig:driving_dynamics_addition}.
This again suggests that conversational dynamics are more strongly shaped by the OP than by the Challenger, aligning with our findings in the main paper.

As shown in this section, we found the results to be both qualitatively and quantitatively similar to those reported in the main paper under stricter conversation length control (restricting to conversations of 4–6 utterances).

\begin{table*}[!ht]
    \centering
    \small
    \begin{tabular}{p{0.2cm}p{1.3cm}p{5cm}p{7cm}}
        \toprule
        {  \#} & {  Category} & {  Dynamics} & {  Examples}\\
        \midrule
        {  1} & {  Tone} & {  \textbf{negative politeness}} & {  SPK1 then concedes, acknowledging a shift in understanding and expressing gratitude.}\\
        {  } & {  } & {  (gratitude, thanks, appreciation)} & {  SPK2 concludes with gratitude and agreement.}\\
        
        {  } & {  } & {  \textbf{collaborative}} & {  The conversation concludes with mutual respect and acknowledgment, with both speakers reflecting on their own viewpoints.}\\
        {  } & {  } & {  (collaborative, shared)} & {  SPK2 concludes by asserting a shared understanding, attempting to resolve the perceived disagreement and establish common ground.}\\

        {  } & {  } & {  \textbf{conciliatory}} & {  SPK1 concedes that SPK2’s explanation has shifted their perspective.}\\
        {  } & {  } & {  (acknowledges, concede)} & {  SPK1 acknowledges their error.}\\
\noalign{\vskip 0.6ex}

        {  } & {  Strategy} & {  \textbf{elaboration}} & {  SPK2 elaborates on their reasoning and acknowledges SPK1’s potential correctness, demonstrating a degree of concession.}\\
        {  } & {  } & {  (elaborate, detailed)} & {  SPK2 offers their own reasons for disliking the episode, providing a detailed explanation.}\\

        {  } & {  } & {  \textbf{agreement}} & {  SPK1 expresses agreement.}\\
        {  } & {  } & {  (agreement)} & {  SPK2 concedes to a point made by SPK1, expressing agreement.}\\

        {  } & {  } & {  \textbf{compromise}} & {  SPK1 responds by agreeing with one aspect of SPK2’s statement while also introducing a contrasting viewpoint.}\\
        {  } & {  } & {  (compromise)} & {  SPK1 concedes a point but expresses a pessimistic view of consumer behavior.}\\

\noalign{\vskip 0.6ex}

        {  } & {  Changes} & {  \textbf{changes in perspective}} & {  SPK1 expresses gratitude for the information, indicating a change in their understanding.}\\
        {  } & {  } & {  (revised, change)} & {  SPK1 acknowledges the potential negative consequences and concedes, changing their view.}\\

        {  } & {  } & {  \textbf{shift to ligher tone}} & {  The tone shifts from inquisitive to reflective and ultimately appreciative.}\\
        {  } & {  } & {  } & {  The tone shifts to a friendly and helpful exchange.}\\
        \bottomrule
    \end{tabular}
    \caption{
    Qualitative analysis of Cluster 1 dynamics from ChangeMyView conversations (selected with less strict length control). 
    Phrases in parentheses are distinguishing words used during the analysis.
    }
    \label{tab:qualitative_ex_cluster1_less_restrict}
\end{table*}
\begin{table*}[!ht]
    \centering
    \small
    \begin{tabular}{p{0.2cm}p{1.3cm}p{5cm}p{7cm}}
        \toprule
        {  \#} & {  Category} & {  Dynamics} & {  Examples}\\
        \midrule
        {  2} & {  Tone} & {  \textbf{dismissive}} & {  SPK2 initiates the conversation with an accusatory and dismissive tone, directly attacking SPK1’s reasoning.}\\
        {  } & {  } & {  (accusatory, dismissive)} & {  SPK2 dismisses the example as irrelevant to their point.}\\
        
        {  } & {  } & {  \textbf{sarcastic}} & {  SPK1 refutes SPK2's claims, employing sarcasm.}\\
        {  } & {  } & {  (sarcasm, sarcastic)} & { SPK1 expresses a sense of resignation, possibly sarcastic.}\\

        {  } & {  } & {  \textbf{defensive}} & {  The tone shifts from informative to defensive.}\\
        {  } & {  } & {  (defensive, refute)} & {  SPK1 responds defensively, attempting to clarify their position and refute SPK2’s interpretation.}\\

        {  } & {  } & {  \textbf{confrontational}} & {  SPK1 accuses SPK2 of sexism and attributes historical disparities to societal constraints.}\\
        {  } & {  } & {  (accuses, blame)} & {  SPK2 starts the conversation with a rhetorical question, implying blame.}\\

\noalign{\vskip 0.6ex}
        {  } & {  Strategy} & {  \textbf{straw man fallacy}} & {  SPK1 accuses SPK2 of using a straw man fallacy.}\\
        {  } & {  } & {  (straw man)} & {  SPK2 suggests a limited perspective and employs a straw man fallacy.}\\

        {  } & {  } & {  \textbf{providing examples}} & {  SPK2 provides examples to support their argument.}\\
        {  } & {  } & {  (examples)} & {  SPK1 rebuts SPK2’s points by dismissing anecdotal evidence.}\\

        {  } & {  } & {  \textbf{analogy}} & {  SPK2 then uses an analogy to challenge SPK1 reasoning.}\\
        {  } & {  } & {  (analogy)} & {  SPK2 then uses an analogy to further clarify their position.}\\

        {  } & {  } & {  \textbf{seeking clarification}} & {  SPK1 responds defensively, seeking clarification.}\\
        {  } & {  } & {  (confusion, seeking clarification)} & {  SPK2 expresses confusion and presses SPK1 to define the specific group to which this obligation applies.}\\

        {  } & {  } & {  \textbf{disagreement}} & {  SPK2 continues to disagree and questions SPK1’s perspective.}\\
        {  } & {  } & {  (disagrees, contrasting)} & {  SPK1 immediately expresses disagreement, employing a comparison to other controversial industries to undermine SPK2 claims.}\\

        {  } & {  } & {  \textbf{direct responses}} & {  SPK1 directly answers the question with disagreement, citing practical concerns. }\\
        {  } & {  } & {  (direct, immediately)} & { }\\
\noalign{\vskip 0.6ex}
        {  } & {  Changes} & {  \textbf{maintains perspective}} & {  SPK1 denies the accusation and reiterates their stance. }\\
        {  } & {  } & {  (reiterates, persists in)} & {  SPK1 persists in their disagreement, providing counter-evidence.}\\

        {  } & { } & {  \textbf{shift to contentious tone}} & { The tone shifts from informative to defensive.}\\
        {  } & {  } & {  } & {  The tone shifts from neutral inquiry to a more challenging and potentially critical stance.}\\
        \bottomrule
    \end{tabular}
    \caption{
    Qualitative analysis of Cluster 2 dynamics from ChangeMyView conversations (selected with less strict length control). 
    Phrases in parentheses are distinguishing words used during the analysis.
    }
    \label{tab:qualitative_ex_cluster2_less_restrict}
\end{table*}

\begin{figure}[t] 
    \centering
    \includegraphics[width=0.48\textwidth]{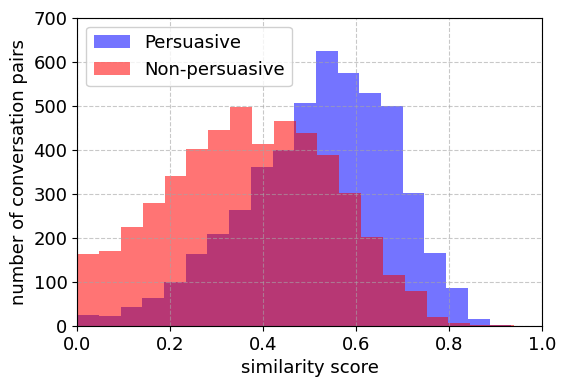}
    \caption{Distribution of similarity scores computed using \measurename for conversations within \groupdelta (\bluetext{blue}) and within \groupnodelta (\redtext{red}).
    }
    \label{fig:within-group-sim}
\end{figure}

\xhdr{Additional result}
\autoref{fig:within-group-sim} demonstrates the intra-group similarity between \groupdelta and \groupnodelta, as described in Section~\ref{sec:intra-group-sim}.

\section{Miscellaneous}
\subsection{Data Anonymization}
We used the CMV dataset, which we accessed through ConvoKit 3.0.1. The dataset includes the usernames of the conversation participants, which we replace with `Speaker1', `Speaker2', and etc. to respect the users' identity, following the procedures outlined in \citet{hua_how_2024}.

\subsection{Implementation Details}
During all generation with Gemini Flash 2.0 via Google Cloud's Vertex AI API, the sampling temperature was set to $0$ for more deterministic
behaviors, and the reported results are from those single runs.
All other settings and parameters were set to the default value.
For gpt-4o models, number of output tokens were limited to $512$.

\subsection{Used Artifacts}
The following is a list of artifacts and their licenses used in the work:

\begin{itemize}

    \item ConvoKit 3.0.1:\\
    \url{https://convokit.cornell.edu/}, MIT License

    \item Gemini Flash 2.0:
    Accessible via Google's Vertex AI API  \url{https://cloud.google.com/vertex-ai?hl=en}
    
    \item gpt-4o-mini-2024-07-18:\\
    a snapshot of gpt-4o-mini from July 18th, 2024. Accessible at a low cost via OpenAI’s API \url{https://platform.openai.com/docs/}

    \item gpt-4o-2024-11-20:\\
    a snapshot of gpt-4o from November 11th, 2024. Accessible via OpenAI's API \url{https://platform.openai.com/docs/}

    \item chatgpt-4o-latest:\\
    most updated version gpt-4o. Accessed on March 2025 via OpenAI's API \url{https://platform.openai.com/docs/}

    \item Sentence Transformers 3.0.0:\\
    \url{https://github.com/UKPLab/sentence-transformers}, Apache License 2.0

    \item DistilBERT base model: \\
    Distilled version of BERT accessed through huggingface API
    \url{https://huggingface.co/distilbert/distilbert-base-uncased}, Apache License 2.0
\end{itemize}

\begin{table*}[!ht]
    \centering
    \small
    \begin{tabular}{p{0.2cm}p{1.3cm}p{5cm}p{7cm}}
        \toprule
        {  \#} & {  Category} & {  Dynamics} & {  Examples}\\
        \midrule
        {  1} & {  Tone} & {  \textbf{vulnerable / seeking validation}} & {  SPK1 reveals vulnerability.}\\
        {  } & {  } & {  (vulnerability, seeks support)} & {  SPK1 expresses vulnerability and seeks support from the others.}\\
        
        {  } & {  } & {  \textbf{defensive / resistant}} & {  SPK1 defensively asserts their confidence.}\\
        {  } & {  } & {  (defensiveness, asserts confidence)} & {  SPK1 expresses defensiveness.}\\
        
        {  } & {  } & {  \textbf{skeptical / doubtful}} & {  SPK2 expresses doubt.}\\
        {  } & {  } & {  (disbelief, doubt)} & {  SPK1 initially expresses disbelief and judgment.}\\
        
\noalign{\vskip 0.6ex}

        {  } & {  Strategy} & {  \textbf{reassurance and support}} & {  SPK2 responds with reassurance and validation, attempting to offer support and build SPK1 confidence.}\\
        {  } & {  } & {  (reassure, validate)} & {  SPK2 uses validation and humor to normalize situation.}\\
        
        {  } & {  } & {  \textbf{avoidance / deflection}} & {  SPK1 deflects with nervous humor and avoidance.}\\
        {  } & {  } & {  (avoid, deflect)} & {  SPK1 uses humor to deflect blame.}\\
        
\noalign{\vskip 0.6ex}

        {  } & {  Changes} & {  \textbf{escalation to conflict}} & {  The tense exchange reveals underlying conflict.}\\
        {  } & {  } & {  (conflict, escalation)} & {  The conversation quickly shifts to conflict and animosity.}\\
        
        {  } & {  } & {  \textbf{shifts in judgment / acceptance}} & {  The initial shock transitions to acceptance.}\\
        {  } & {  } & {  (shift, acceptance)} & {  SPK1 initially expresses disbelief and judgment.}\\
        
        \midrule
        {  2} & {  Tone} & {  \textbf{playful / teasing}} & {  The conversation begins with playful banter and lighthearted teasing, establishing a jovial mood.}\\
        {  } & {  } & {  (playful, teasing)} & {  SPK1 uses teasing to express annoyance.}\\
        
        {  } & {  } & {  \textbf{lighthearted / humorous}} & {  The conversation begins with a lighthearted exchange, marked by playful teasing and inside jokes.}\\
        {  } & {  } & {  (lighthearted, humor, jokes)} & {  Initial warmth shifts to awkwardness as one speaker attempt at humor falls flat.}\\
        
\noalign{\vskip 0.6ex}

        {  } & {  Strategy} & {  \textbf{rapport through humor}} & {  SPK2 uses humor to downplay SPK3 concerns.}\\
        {  } & {  } & {  (jokes, humor, playful)} & {  The banter and disagreement are fueled by inside jokes and shared humor.}\\
        
        {  } & {  } & {  \textbf{playful negotiation / competition}} & {  Playful negotiation occurs.}\\
        {  } & {  } & {  (playful challenge)} & {  The conversation shifts to a competitive dynamic as two speakers vie for attention and affection.}\\
        
\noalign{\vskip 0.6ex}

        {  } & {  Changes} & {  \textbf{resolution / reconciliation}} & {  The conversation concludes with a reconciliation and a renewed sense of connection between SPK2 and SPK4.}\\
        {  } & {  } & {  (resolution, reconciliation)} & {  The conversation concludes with a display of friendship and mutual support between two speakers.}\\

        \bottomrule
    \end{tabular}
    \caption{
    Qualitative analysis of the two identified clusters from the \textit{Friends} dialogues. 
    Dynamics found in cluster 1 are on the top; 
    dynamics found in cluster 2 are on the bottom. 
    Phrases in parentheses are distinguishing words used during the analysis.
    }
    \label{tab:qualitative_ex_friends}
\end{table*}

\begin{table*}[!ht]
    \centering
    \small
    \begin{tabular}{p{0.2cm}p{1.3cm}p{5cm}p{7cm}}
        \toprule
        {  \#} & {  Category} & {  Dynamics} & {  Examples}\\
        \midrule
        {  1} & {  Tone} & {  \textbf{polite / appreciative}} & {  SPK2 responds with gratitude.}\\
        {  } & {  } & {  (gratitude, appreciation, thanks)} & {  SPK2 acknowledges the information and expresses gratitude.}\\
        
        {  } & {  } & {  \textbf{gentle corrective}} & {  SPK1 politely informs SPK2 about a mistake they made.}\\
        {  } & {  } & {  (polite correction, guidance)} & {  SPK1 gently corrects SPK2, providing further resources and maintaining a polite tone.}\\
        
        {  } & {  } & {  \textbf{supportive / reassuring}} & {  SPK2 responds with reassurance and encouragement.}\\
        {  } & {  } & {  (support, reassurance, encouragement)} & {  SPK3 offers supportive feedback.}\\

\noalign{\vskip 0.6ex}

        {  } & {  Strategy} & {  \textbf{repetition for emphasis}} & {  SPK2 repeats the same link again.}\\
        {  } & {  } & {  (repeats, persistence)} & {  SPK1 repeats the request for multiple images.}\\
        
        {  } & {  } & {  \textbf{evidence / justification}} & {  SPK1 provides evidence to support their claim.}\\
        {  } & {  } & {  (evidence, rationale)} & {  SPK2 defends the new category by providing a rationale.}\\
        
        {  } & {  } & {  \textbf{polite mitigation}} & {  SPK1 initiates the conversation with a polite request.}\\
        {  } & {  } & {  (polite request)} & {  SPK1 politely requests a change in SPK2 behavior, providing a rationale.}\\
        
\noalign{\vskip 0.6ex}

        {  } & {  Changes} & {  \textbf{ conceding }} & {  SPK2 later concedes and expresses willingness to proceed.}\\
        {  } & {  } & { (concede)} & {  SPK2 concedes and expresses a willingness to proceed, ending the conversation on a cooperative note.}\\
        
        \midrule
        {  2} & {  Tone} & {  \textbf{defensive}} & {  SPK2 responds defensively, justifying their actions.}\\
        {  } & {  } & {  (defensive, blame)} & { SPK2 responds defensively, offering an explanation and shifting blame to another source.}\\

        {  } & {  } & {  \textbf{sarcastic / dismissive}} & {  SPK3 uses sarcasm, expressing frustration.}\\
        {  } & {  } & {  (sarcasm, dismiss)} & {  SPK1 initiates with a sarcastic tone, questioning another user’s actions.}\\
        
        {  } & {  } & {  \textbf{accusatory / confrontational}} & {  SPK1 initiates the conversation with an accusatory tone, suggesting an edit war.}\\
        {  } & {  } & {  (accuses)} & {  SPK3 accuses another user of disregarding established protocols and imposing their view unilaterally.}\\
\noalign{\vskip 0.6ex}

        {  } & {  Strategy} & {  \textbf{rebuttal / counter-argument}} & {  SPK2 immediately rebuts SPK1 assertion.}\\
        {  } & {  } & {  (rebut, counter)} & {  SPK3 counters with examples and sarcasm, expressing frustration.}\\
        
        {  } & {  } & {  \textbf{appeals to policy / guidelines}} & {  SPK1 appeals to a guideline.}\\
        {  } & {  } & {  (policy, guideline)} & {  SPK2 defends their position by quoting policy.}\\
        
        {  } & {  } & {  \textbf{accusations / fallacy claims}} & {  SPK3 attempts to identify fallacies in SPK4 reasoning.}\\
        {  } & {  } & {  (fallacy, accusation)} & {  SPK3 echoes SPK1 accusation, suggesting continued doubt or disagreement with SPK2 defense.}\\
        
\noalign{\vskip 0.6ex}

        {  } & {  Changes} & {  \textbf{persistent disagreement} } & {  SPK1 reiterates their original point with persistent disagreement.}\\
        {  } & {  } & {  (persistent, reiterate)} & {  The conversation involves persistent disagreement and defense of positions.}\\
        
        {  } & {  } & {  \textbf{unresolved / rigidity}} & {  SPK2 repeats SPK1 refutations verbatim.}\\
        {  } & {  } & {  (unresolved)} & {  Conversation appears unresolved, with SPK2 not responding to detailed explanation.}\\

        \bottomrule
    \end{tabular}
    \caption{
    Qualitative analysis of the two identified clusters from German Wikipedia talk-page conversations. 
    Dynamics found in cluster 1 are on the top; dynamics found in cluster 2 are on the bottom. 
    Phrases in parentheses are distinguishing words used during the analysis.
    }
    \label{tab:qualitative_ex_wiki}
\end{table*}

\end{document}